\documentclass[preprint,12pt]{elsarticle}

\usepackage{amssymb}
\usepackage{amsmath}
\usepackage{lineno}
\linenumbers

\usepackage{amsmath,amsfonts}
\usepackage{algorithmic}
\usepackage{algorithm}
\usepackage{array}
\usepackage[caption=false,font=normalsize,labelfont=sf,textfont=sf]{subfig}
\usepackage{textcomp}
\usepackage{stfloats}
\usepackage{url}
\usepackage{verbatim}
\usepackage{graphicx}
\usepackage[utf8]{inputenc} %
\usepackage[T1]{fontenc}    %
\usepackage{hyperref}       %
\usepackage{url}            %
\usepackage{booktabs}       %
\usepackage{amsfonts}       %
\usepackage{nicefrac}       %
\usepackage{microtype}      %
\usepackage{xcolor}         %
\usepackage{graphicx}
\usepackage{tabularx}
\usepackage{listings}
\usepackage{placeins} %
\usepackage{longtable}
\usepackage{multirow}
\usepackage{makecell}

\usepackage{enumitem}

\usepackage{pifont}
\newcommand{\cmark}{\ding{51}}%
\newcommand{\xmark}{\ding{55}} %

\definecolor{bluetext}{RGB}{72, 120, 209}
\definecolor{orangetext}{RGB}{239, 134, 74}
\definecolor{pptgreen}{RGB}{47,190,112}
\definecolor{pptred}{RGB}{194,11,10}
\definecolor{sred}{RGB}{212,17,89}
\definecolor{sblue}{RGB}{0,90,181}

\newcommand{\rebuttal}[1]{\textcolor{black}{{#1}}}

\newcommand{\revision}[1]{\textcolor{black}{{#1}}}
\newcommand{\revisionred}[1]{\textcolor{black}{{#1}}}

\usepackage{arydshln}

\usepackage{layout}

\journal{Medical Image Analysis}

\begin{document}
\nolinenumbers %
\begin{frontmatter}

\title{WBCAtt+: Fine-Grained Pixel-Level Morphological Annotations for White Blood Cell Images}

\affiliation[label1]{organization={Rapid-Rich Object Search (ROSE) Lab, School of Electrical and Electronic Engineering, Nanyang Technological University},
            country={Singapore}}

\affiliation[label2]{organization={Shanghai University of Finance and Economics},
            city={Shanghai},
            country={China}}

\author[label1]{Satoshi Tsutsui\corref{equal}}
\author[label1]{Winnie Pang\corref{equal}}
\author[label2]{Shuting He}
\author[label1]{Bihan Wen\corref{cor1}}

\cortext[equal]{Equal Contribution.}
\cortext[cor1]{Corresponding author: Bihan Wen (email: bihan.wen@ntu.edu.sg)}

\begin{abstract}
The microscopic examination of white blood cells (WBCs) plays a fundamental role in pathology and is essential for diagnosing blood disorders such as leukemia and anemia. To support further research on WBC images, multiple datasets have been proposed. However, they mainly annotate cell categories, and lack detailed morphological characteristics that pathologists use to explain their interpretations of cells. To address this gap, we introduce WBCAtt+, a novel dataset of WBC images densely annotated with 11 morphological attributes and five pixel-level cell components. With 113k image-level labels and 10k segmentation maps, WBCAtt+ is the first to provide comprehensive annotations for WBC images. Leveraging this dataset, we provide baseline models for attribute recognition and semantic segmentation. We also design an attribute recognition model to incorporate compositional structure of cells, further improving the recognition performance. Lastly, we showcase various applications enabled by our dataset, such as explainable AI models, including counterfactual example generation. \revision{The dataset and code are publicly available\footnote{\url{https://doi.org/10.57967/hf/8143}}}.
\end{abstract}


\end{frontmatter}
\let\thefootnote\relax\footnotetext{Preprint accepted to \textit{Medical Image Analysis} on May 19, 2026.}

\begin{figure*}[!t]
  \centering
  \includegraphics[trim=0 607pt 0 0, clip, width=\linewidth]{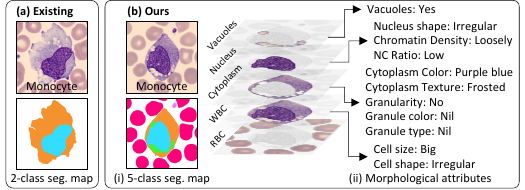}
  \caption{Our WBCAtt+ dataset provides (i) fine-grained segmentation maps with (ii) detailed morphological attributes, making it richer than existing datasets.}
  \label{fig:overview}
\end{figure*}

\section{Introduction}
The microscopic examination of human blood samples provides valuable insights into pathology and hematology. For example, recognizing white blood cells (WBCs) is a fundamental part of diagnostic methods~\cite{Blumenreich1990TheWB}. WBC recognition has been a subject of machine learning research since at least the 1990s~\cite{song1997incorporating} and has also been widely studied within the medical image computing community~\cite{zhang2020corruption,vigueras2021parallel,salehi2022unsupervised}. To facilitate further research, previous work~\cite{labati2011all, Mohamed2012AnET, acevedo2020pbc, kouzehkanan2022large, rezatofighi2011automatic} has released annotated datasets of cell images categorized by cell type.

However, these efforts typically prioritize cell type identification without considering the fine-grained structures essential for model explainability. Existing datasets often fail to provide annotations at the level of detail that pathologists use for their own cell recognition. For instance, holes in the cytoplasm are called cytoplasmic vacuoles \revision{(hereafter referred to as vacuoles)}, as depicted in Figure~\ref{fig:attributes} (t) vs. (u). Vacuoles are reported as discriminative features for monocytes and neutrophils~\cite{dale2008phagocytes}, and are reported to indicate certain disorders~\cite{van2001peripheral}. Unfortunately, current datasets do not provide annotations to segment vacuoles, nor do they include a label to indicate the existence of holes in the cytoplasm. Generally, existing datasets lack morphological characteristics, including vacuoles, which are key underlying factors in how clinicians recognize WBCs. This lack of detailed morphological annotations limits the development of explainable artificial intelligence (XAI), which is crucial in healthcare, where clear explanations of a model’s conclusions are necessary. Therefore, while such detailed annotations may have been limited due to their labor-intensive nature, we believe it is worth investing in them to capture the same detailed observations that clinicians do.

This paper introduces WBCAtt+ (Figure~\ref{fig:overview}): a novel peripheral blood cell image dataset annotated with morphological attributes and pixel-level details that pathologists recognize. \revision{The innovation of this work lies in providing a large-scale WBC dataset that integrates fine-grained morphological attributes with pixel-level segmentation, with key differences from existing datasets summarized in Table~\ref{tab:dataset_info}}. Each cell image, sourced from the PBC dataset \cite{acevedo2020pbc}, is annotated with 11 attributes and a semantic segmentation map. We annotated a total of 10,298 WBC images, resulting in 113k labels (11 attributes × 10.3k images) and 10.3k semantic segmentation maps. To the best of our knowledge, this is the first public dataset to include such detailed attributes. For segmentation, while we are not the first, our dataset is more than 10 times larger than prior work, and contains more semantic classes.

\begin{table}[!t]
\centering
\caption{Comparison of Existing WBC Attribute and Segmentation Datasets.}
\resizebox{\linewidth}{!}{ %
\footnotesize
\begin{tabular}{cccccccc}
\toprule
\multirow{2}{*}{ Dataset} & \multirow{2}{*}{\begin{tabular}[c]{@{}c@{}}\revision{\# Unique}\\ \revision{Cell Images}\end{tabular}} & \multirow{2}{*}{\begin{tabular}[c]{@{}c@{}}\revision{\# Morphological}\\ \revision{Attributes}\end{tabular}} & \multicolumn{5}{c}{\revision{Segmentation Class}} \\  \cmidrule{4-8}
 &  &  & Neucleus & Cytoplasm & Vacuoles & Platelets & RBC \\
 \midrule
\revision{LISC~\cite{rezatofighi2011automatic}} & \revision{250} & \revision{\xmark} & \revision{\cmark} & \revision{\cmark} &  &  &  \\
Zheng et al.~\cite{zheng2018} & 400 & \xmark & \cmark & \cmark &  &  &  \\
RaabinWBC~\cite{kouzehkanan2022large}-Seg & 1,145 & \xmark & \cmark & \cmark &  &  &  \\
\revision{LeukemiaAttri~\cite{rehman2024large}} & \revision{7,857} & \revision{7} &  &  &  &  &  \\
\textbf{WBCAtt+} (Ours) & 10,298 & 11 & \cmark & \cmark & \cmark & \cmark & \cmark \\
\bottomrule
\end{tabular}
}
\label{tab:dataset_info}
\end{table}

Based on the proposed dataset, we train baseline models to perform attribute recognition (Sec.~\ref{sec:attprediction}) and semantic segmentation (Sec.~\ref{sec:segmentation}). Furthermore, we develop a model to use the segmentation maps to better recognize attributes. The model design is inspired by the recognition process of pathologists who naturally distinguish different cell components (nucleus, cytoplasm, vacuoles, etc.) to recognize the attributes.  We demonstrate that this model, enabled by our new dataset, can significantly improve attribute recognition performance compared to the baselines (Sec.~\ref{sec:segrec_exp}). Last but not least, we believe that our dataset enables various new applications, such as automated cell analysis. To demonstrate this, we showcase specific applications that can be developed using our newly-introduced dataset (Sec.~\ref{sec:applications}).

\vspace{5pt}
\noindent \textbf{Contributions}. In summary, this paper makes the following contributions. 
\begin{enumerate}[itemsep=0pt, parsep=0pt, topsep=0pt]
\item We annotated morphological attributes of WBCs for the first time, along with much larger-scale and finer-grained semantic segmentation maps than existing work.
\item We provide baseline models for attribute recognition and semantic segmentation.
\item We demonstrate how our new dataset enables the development of a model that leverages semantic segmentation to improve attribute recognition performance.
\item We demonstrate various applications enabled by our new dataset.
\end{enumerate}
\vspace{10pt}

This work is an extension of our previous NeurIPS paper~\cite{tsutsui2023wbcatt}, with several improvements and two entirely new contents. First, we have annotated 10k segmentation maps and trained semantic segmentation models (Sec.~\ref{sec:seg-anno} and \ref{sec:segmentation}). Second, we design a model that effectively uses the knowledge of cell structure to better recognize cell attributes~(Sec.~\ref{sec:segrec_model}). \revision{A side-by-side comparison with the conference version is summarized in Appendix Table~\ref{tab:journal-update}}.

\section{Related Work}
Prior work exists on cell classification~\cite{song1997incorporating,labati2011all, Mohamed2012AnET, acevedo2020pbc,rezatofighi2011automatic,chen2021transmixnet, chen2022accurate,yan2021development} and segmentation~\cite{kouzehkanan2022large, zheng2018,li2022segmentation,rao2023effective}. However, the datasets used in prior work do not capture the details that pathologists recognize. \revision{Table~\ref{tab:dataset_info} summarizes the differences between our dataset and other attribute and segmentation datasets for WBCs.} Segmentation maps have smaller scales and fewer classes, and no prior datasets annotate morphological attributes, which are essential explanatory factors when pathologists recognize cells. Our work aims to address this overlooked aspect. We note that the cell attribute dataset by Rehman et al.~\cite{rehman2024large}, which cites our conference version~\cite{tsutsui2023wbcatt}, was published after our work and therefore does not affect our claim of being the first to annotate morphological attributes.

Beyond WBCs, attribute datasets with segmentation masks have played a pivotal role in computer vision, inspiring many applications and fostering methodological advances. The prominent example is the CUB Bird dataset~\cite{wahcub_200_2011} that provides fine-grained attributes for bird classification as well as bird segmentation masks. This has been a good test bed for metric learning~\cite{mishra2021effectively}, few-shot learning~\cite{xian2018zero}, few-shot segmentation~\cite{saha2022improving}, and explainable AI~\cite{koh2020concept}. Attributes are often considered key interpretable elements in computer vision systems. For example, interpretable autonomous driving systems can be developed using explainable attributes~\cite{xu2020explainable}. In medical domains, X-ray images~\cite{quesada2022mtneuro, Wu2021ChestID} and skin disease images~\cite{codella2018skin,daneshjou2022skincon} have been annotated with attributes. They share a similar motivation with our work, although only Codella et al.~\cite{codella2018skin} provide segmentation annotations \revision{for the skin images}. 

\section{Methodology}\label{sec:dataset}

\begin{figure*}[!tb]
    \centering
    \includegraphics[trim=0 28pt 0 0, clip, width=0.8\textwidth]{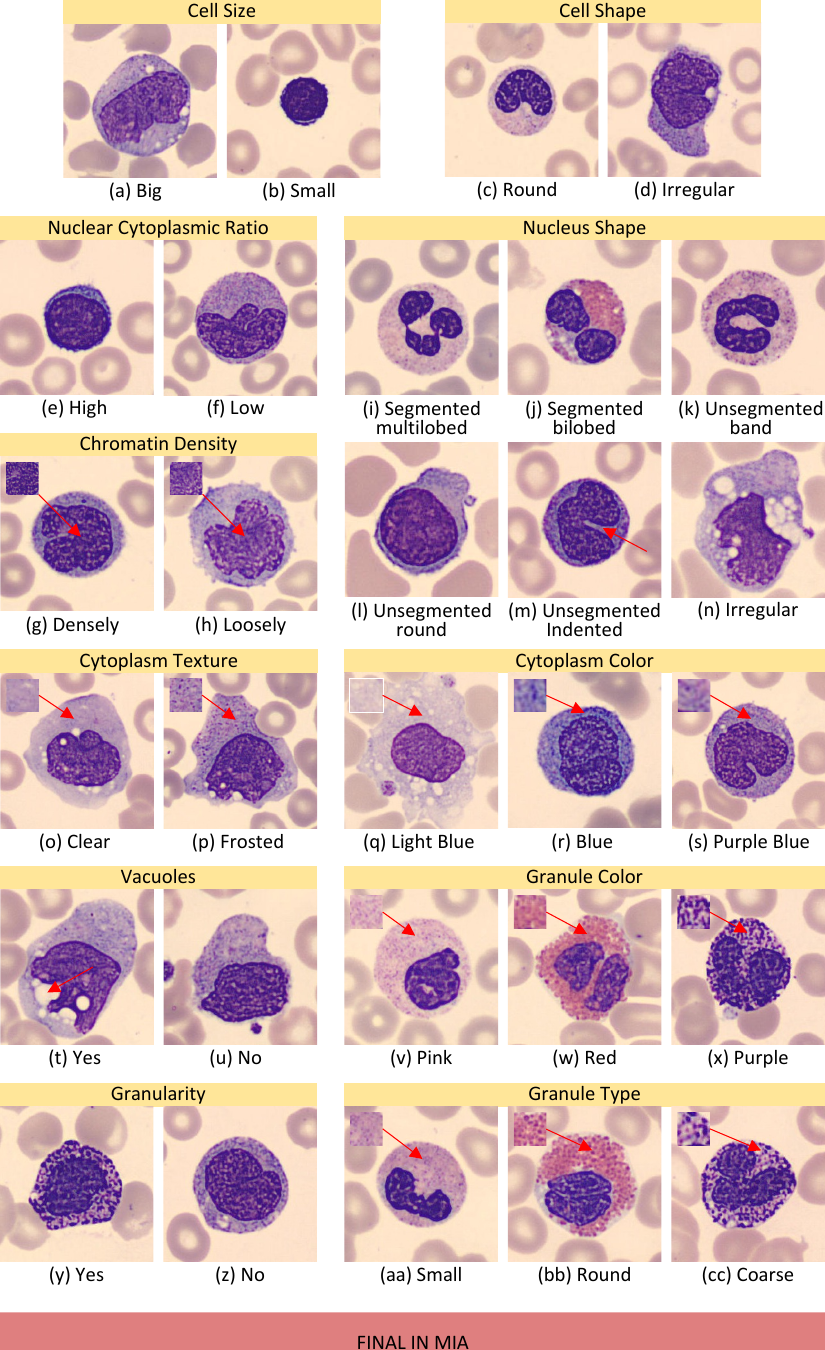}
    \caption{Sample images of each morphological attribute, which plays a key role in recognizing cell types. Table~\ref{tab:attributes_def} provides a description of each attribute.}\label{fig:attributes}
\end{figure*}

\begin{table*}[!tb]
    \centering
    \caption{Attributes Differentiating Types of White Blood Cells}
    \label{tab:attributes_def}
    {\footnotesize	
    \begin{tabularx}{\textwidth}{lX}
        \toprule
        Cell Size & The diameter of a WBC. Cells are classified as big if their diameter is more than twice that of red blood cells.  \\
        Cell Shape & The shape of WBCs, which can be round or irregular, providing clues about the cell type and its interactions. \\
        Nucleus Shape & The nucleus shape varies, e.g., segmented into multiple lobes or not. \\
        NC Ratio & The ratio of the nucleus to the cytoplasm, with a high ratio typically indicating lymphocytes. \\
        Chromatin Density & The compactness of chromatin in the nucleus. Lymphocytes typically have dense chromatin, while monocytes have a looser structure.\\
        Cytoplasm Color & Reveals cell type and maturation based on color variations ranging from light blue to purple-blue. \\
        Cytoplasm Texture & Variations in texture, including a frosted appearance due to tiny, dust-like granules or a clear appearance. \\
        \revision{Vacuoles}  & Small cavities in the cytoplasm, primarily found in monocytes and neutrophils. \\
        Granularity & The presence of prominent stainable granules in the cytoplasm that distinguish between granulocytes and agranulocytes. \\
        Granule Color & Differentiates granulocytes based on color, with pink for neutrophils, red for eosinophils, and purple for basophils. \\
        Granule Type & Characterizes the granules in granulocytes, with each type having unique granules reflecting their roles in immunity. \\
        \bottomrule
    \end{tabularx}
    }
\end{table*}

\begin{figure*}[!tb]
  \centering
  \includegraphics[trim=0 551pt 0 0, clip, width=0.8\textwidth]{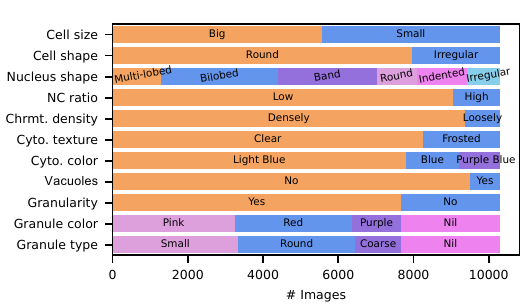}
  \caption{The distribution of values per attribute. The distribution represents the results of annotating all typical WBCs from the PBC dataset, which is the image source we utilized. We did not actively control or manipulate the distribution. See Table~\ref{tab:attributes_def} and Figure~\ref{fig:attributes} for their definition and example images. Appendix Table~\ref{tab:attributes} has the extract numbers of this plot.}
  \label{fig:distribution}
\end{figure*}

\begin{figure*}[!tb]
  \centering
  \includegraphics[width=\linewidth]{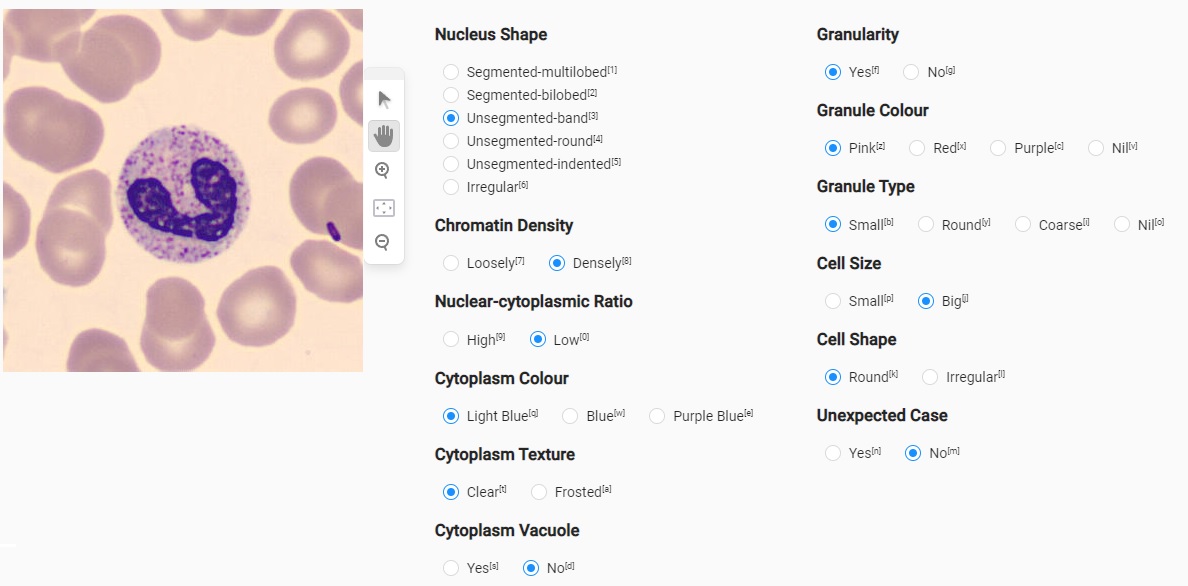}
    \caption{The screenshot of Label Studio, the labeling tool used to annotate the attribute of the WBCAtt+ dataset. Annotators can determine the cell type from the name associated with the cell, which is provided before image selection. The image to be annotated is presented in a clear and organized manner, enabling annotators to easily navigate and zoom in on the image. Attribute selections are organized using radio buttons, ensuring that only one choice can be selected at a time. All attributes must be selected before submitting and proceeding to the next image.}
  \label{fig:annotation-interface}
\end{figure*}

\subsection{Attribute Annotation}\label{sec:attribute-annotation-main}
We identified 11 cell attributes through discussions with pathologists. These attributes are summarized in Table~\ref{tab:attributes_def} and Figure~\ref{fig:attributes}. To annotate these attributes, we utilized the PBC dataset~\cite{acevedo2020pbc}, which provide normal and healthy blood cell images collected at the Hospital Clinic of Barcelona. It includes five typical WBCs, comprising 1,218 basophils, 3,117 eosinophils, 1,420 monocytes, 3,329 neutrophils, and 1,214 lymphocytes. We annotated 11 attributes for these 10,298 images, resulting in 113,278 image-attribute pairs. Figure~\ref{fig:distribution} visualizes the distribution of the attribute annotations.

\textbf{Attribute Definition Process}. Since there was no formally established set of attributes or ontology, we initiated our work with discussions with pathologists working in a laboratory at a healthcare company that develops digital cell imaging analyzers~\cite{sysmex2023di60}, who gave us initial attributes (shown in Appendix Table~\ref{tab:initial-att}). 

We then reviewed relevant textbooks and research papers focusing on the morphological characteristics of WBCs. Subsequently, we refined the attribute set through further discussions with the pathologists and by manually inspecting a thousand WBC images. This process yielded a total of 11 attributes, each supported by at least one medical literature reference (see ~\ref{sec:att_def}). While the samples we inspected were the five major WBC types from healthy individuals, we expect the resulting attributes to sufficiently describe the morphological characteristics that may emerge in response to certain diseases,  as demonstrated with cell images of Covid-19 patients in Sec.~\ref{sec:cell-morph-analyzer}. 

\textbf{Annotation with Quality Control}. To ensure reliable annotations across over 10k images, we devised a rigorous, iterative process involving pathologists, research scientists, and biomedical students who had strong knowledge of cell structures. In the initial stage, the students annotated the images, while being informed of the specific WBC category for each image. Following this, our research scientists meticulously examined each image and its assigned attributes, meaning that every image was inspected by at least two individuals before assigning the final label. When ambiguities arose, we discussed with the pathologists who defined the attributes with us, ensuring a consensus on labeling. We engaged three pathologists, each with over 10 years of experience in the relevant domain. Their work involves identifying and analyzing white blood cell types in microscopic images, ensuring their expertise is directly aligned with the annotation task. Our annotation interface is shown in Figure~\ref{fig:annotation-interface}. Further details on the quality control are in the \ref{sec:annnotation}.

\begin{table*}[!tb]
    \centering
    \revision{
    \caption{Agreement (\%) of Different Annotators for Attribute Annotation.}}
    \label{tab:agreement-per-attribute}
    \resizebox{\textwidth}{!}{%
    \revision{
    \begin{tabular}{lccccccccccc}
    \toprule
     & Cell Size & Cell Shape & Nucleus Shape & NC Ratio & Chromatin Density & Cyto. Texture & Cyto. Color & Vacuoles & Granularity & Granule Color & Granule Type \\
    \midrule
    Agree Rate & 89.80 & 95.80 & 90.30 & 97.60 & 96.50 & 94.90 & 95.10 & 98.60 & 99.90 & 98.70 & 99.70 \\
    Kappa~\cite{mchugh2012interrater} & 79.21 & 88.68 & 88.22 & 92.53 & 86.12 & 86.87 & 90.59 & 92.22 & 99.79 & 98.19 & 99.58 \\
    \bottomrule
    \end{tabular}
    }}
\end{table*}

\textbf{Label Quality}. We assessed the reliability of our annotation process by replicating it on a subset of 1,000 images with different annotators. Out of the 11,000 attribute annotations, 10,569 were consistent with the original annotations, giving an agreement rate of approximately $10569/11000 \approx $ 96.1\%. As a reference, a similar dataset on a skin disease reports an agreement rate of 92\% to 97\%~\cite{daneshjou2022skincon}, indicating that our dataset is of comparable quality to existing datasets. \revision{Per-attribute agreement rates are reported in Table 3 %
(second row). Furthermore, we compute the kappa statistic~\cite{mchugh2012interrater} as a statistical measure that accounts for agreement occurring by chance and is widely used for categorical annotation tasks. As shown in Table 3 %
(third row), most attributes exhibit high agreement ($0.81 \leq \kappa \leq 1.00$), which is interpreted as ``strong to almost perfect agreement''~\cite{mchugh2012interrater}. Only the cell size attribute shows slightly lower agreement (79.21). This is likely due to the subjective nature of the size annotation task, where annotators must make a binary judgment (large vs.\ small) based on visual comparison with red blood cells, leading to increased variability on borderline cases.}

\revision{
\textbf{Clinical Interpretability and Validation}.We computed the correlation between each binarized attribute and WBC class, with the resulting matrix presented in Appendix Figure~\ref{fig:att-correlation}. Overall, these correlations align well with clinical knowledge. For instance, lymphocytes show the two highest correlations ($0.9084$ and $0.8036$) with a high NC ratio and a round nucleus shape, respectively, compared to other attributes. Monocytes exhibit a high correlation ($0.7328$) with an indented nucleus shape. These attributes are recognized as key clinical discriminants for classifying lymphocytes versus monocytes~\cite{bain2017blood}. For granulocytes (Basophils, Eosinophils, and Neutrophils), the attributes most strongly  correlated with class are granule type and granule color, consistent with descriptions reported in the medical literature~\cite{azwai2007morphological, tigner2020histology, carr2021clinical}. 
}

\subsection{Segmentation Annotation}\label{sec:seg-anno}
\begin{figure}[t!]
  \centering
  \includegraphics[width=0.8\linewidth]{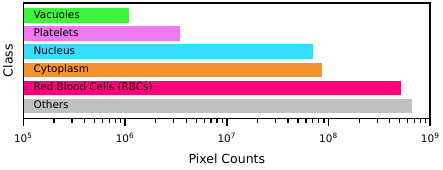}
  \caption{The distribution of semantic classes in pixels. The distribution is imbalanced, with vacuoles having the lowest number of pixels, making them challenging to segment.}
  \label{fig:seg-distribution}
\end{figure}

\begin{figure}[t!]
  \centering
  \includegraphics[width=0.9\linewidth]{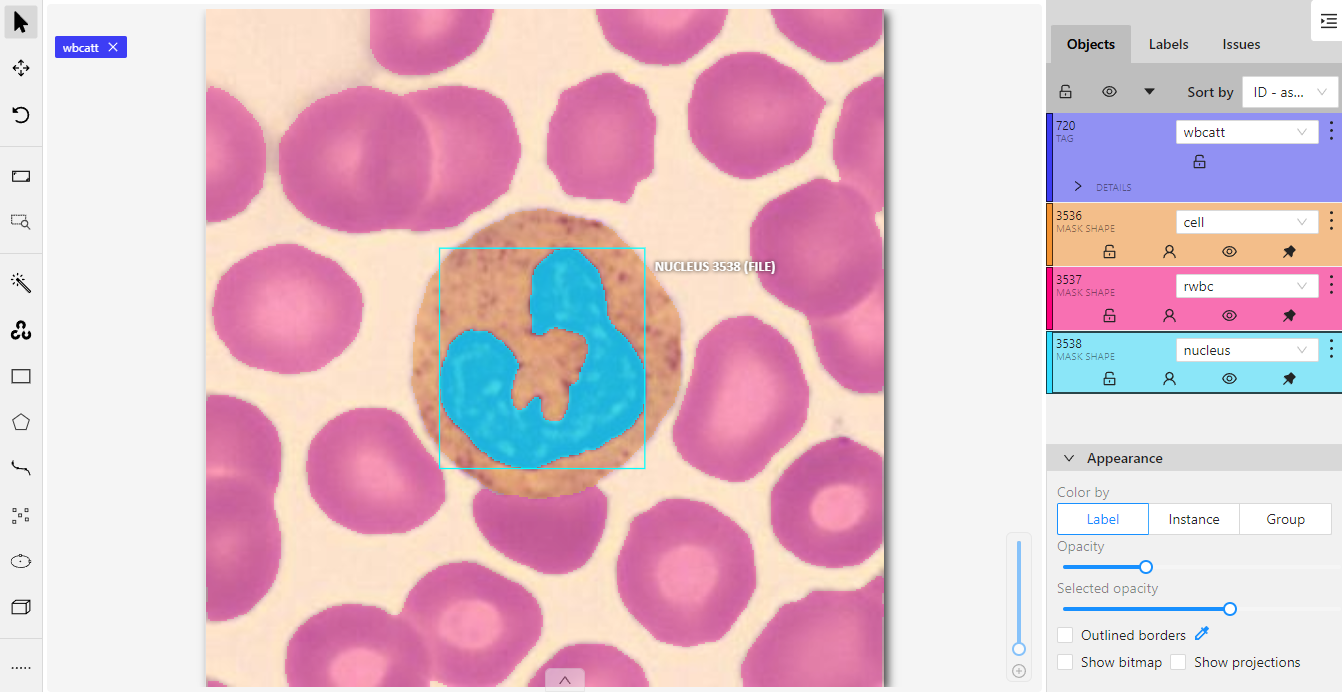}
    \caption{Screenshot of CVAT, the annotation tool used for annotating segmentation maps of the WBCAtt+ dataset. The dropdown list under `wbcatt' allows annotators to reference the attributes for each image, along with the WBC class of the image. }
  \label{fig:annotation-interface-seg}
\end{figure}

We annotated semantic segmentation maps for each of the 10,298 images, with examples illustrated in Figure~\ref{fig:sampleseg} (a)-(j).  Unlike attribute annotations where no prior work was available, prior work annotated two semantic segmentation classes~\cite{kouzehkanan2022large, zheng2018}: nucleus and cytoplasm. While they are important, the cell images also contain other significant objects such as vacuoles~\cite{dale2008phagocytes} and platelets~\cite{al2021improving}. To include these, we define five semantic classes: red blood cells (RBCs), cytoplasm, nucleus, platelets, and vacuoles. Figure~\ref{fig:seg-distribution} and Figure~\ref{fig:annotation-interface-seg} show the distribution of pixels across the dataset and the annotation interface on an image, respectively. Vacuoles and platelets are very small and have much fewer pixels than others, making the segmentation task challenging.

\textbf{Two Stage Annotation}. As pixel-wise annotation is significantly more time-consuming than image-wise annotation, we adopted a two-step approach to annotate them efficiently. Initially, we manually annotated the segmentation masks for 2000 images. Subsequently, we utilized these annotated images to train a CNN-based semantic segmentation for semi-automatic annotation. Using the trained model, we generated segmentation masks for the remaining images, which are manually corrected by human annotators. 

\textbf{Mask Annotation with Quality Control}. To ensure reliable segmentation annotations, we implemented a rigorous quality control process similar to attribute annotation. We hired four experienced annotators with biomedical backgrounds to refine and correct the generated segmentation masks. Each annotator participated in a briefing session to familiarize themselves with the annotation process and establish a common baseline agreement. A screenshot of the materials provided in the session is shown in Appendix Figure~\ref{fig:mask_guide}. After this refinement process, our research scientist meticulously reviewed each image to ensure mask consistency and minimize errors. In this way, each image was being checked by at least two individuals before finalizing the segmentation map. Further details are available in \ref{sec:annotation-seg}.

\textbf{Mask Quality}. To estimate the reliability of our annotation procedure, we randomly selected 2000 images and replicated the annotation process with different annotators. \revision{The kappa statistic~\cite{mchugh2012interrater} between the two sets of annotations is 97.48\%, indicating a high level of agreement.} Moreover, we calculated the IoU between pairs of masks corrected by different annotators. The resulting mIoU for these pairs was 94.64\%. For reference, previous studies in generic segmentation have reported inter-annotator consistency ranging from 85\% to 91\%~\cite{gupta2019lvis,kuznetsova2020open}. Therefore, the quality of our segmentation annotation is comparable to other datasets. 

\subsection{Cell Structure-Aware Model}\label{sec:segrec_model}
In this section, we discuss how our segmentation annotations can serve as a valuable resource for improving cell attribute recognition by enabling a model inspired by pathologists’ recognition process.
Pathologists naturally distinguish compositional cell structures, such as red blood cells, nucleus, and cytoplasm, as part of their process for recognizing cell attributes. Inspired by this, we hypothesize that rooting the attribute recognition process with cell structure can also improve the performance of computational models. To verify this, we design a model to utilize cell structure to recognize the cell attributes. Specifically, we develop a model that can recognize attributes in a compositional manner -- segmenting the cells into parts and then recognizing attributes by aggregating the representation of each part. 

This compositional cell parsing can be further explained by theories of human vision and computational signal processing. Human vision~\cite{marr1982vision,egan_orlandi_2010} is modular, processing distinct aspects of visual information in parallel before integrating them. Inspired by this, our model processes pixels from the same semantic class (e.g., nucleus, cytoplasm) separately and then aggregates them, reflecting this modular structure. Additionally, processing per semantic class can be viewed as a form of non-local operator~\cite{buades2005non}, where the context of the entire image is captured based on pixelwise similarity. By grouping pixels within the same class, we preserve the cell's structural context, enabling more accurate recognition of cellular attributes. 

\begin{figure*}[!tb]
  \centering
  \includegraphics[trim=0 483pt 0 0, clip, width=0.9\linewidth]{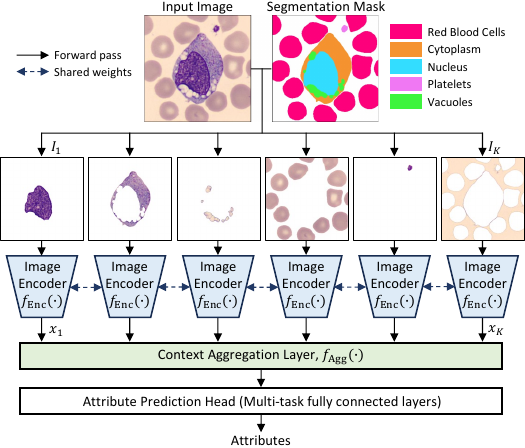}
  \caption{
  We design an attribute recognition model that segments cell images into constituents, followed by attribute recognition. Our model encodes segmented images into feature vectors, which are then aggregated into a comprehensive image representation. We explored several context aggregation mechanisms such as max-pooling or dot-product attention. The attribute prediction head utilizes the aggregated representation to make final attribute predictions.
  }
  \label{fig:partmodel}
\end{figure*}

\textbf{Model Design}: The network is illustrated in Figure~\ref{fig:partmodel}. It takes the input of $K$ segmented images $\{I_i\}_{i=1}^{K}$ of a cell, where $I_i$ represents a cell image masked with the $i$-th semantic segmentation class. Each $I_i$ is then passed through a shared image encoder $f_{\text{Enc}}(\cdot)$ to extract a feature vector $x_i = f_{\text{Enc}}(I_i) \in \mathbb{R}^d$. Subsequently, the network aggregates these segmentation-based image representations  $\{x_i\}_{i=1}^{K} \in \mathbb{R}^{Kd}$ using a context aggregation layer. The aggregation function $f_{\text{Agg}}(\cdot)$ can vary from simple average to attention~\cite{vaswani2017attention} (i.e., average whose weights are trainable and computed per image). We explore five variants, as summarized in Table~\ref{tab:aggregation}. Following the aggregation process, the resulting output is used as the image representation for the multi-task attribute recognition head. 

\section{Experiments}\label{sec:exp}
This section presents empirical results on our dataset. Throughout the section, our data split consists of 6,169 training images, 1,030 validation images, and 3,099 test images, with cell type distributions maintained across all sets. The exact split is available as part of the dataset files.

\subsection{Cell Segmentation}\label{sec:segmentation}
This section empirically examines the performance of existing semantic segmentation models in predicting segmentation maps from the cell images in our dataset.

\textbf{Baselines}. We employ widely-used models for semantic segmentation in computer vision, namely fully-convolutional network (FCN~\cite{long2015fully}) variants~\cite{xiao2018unified}. We also include SegFormer~\cite{xie2021segformer}, a model based on vision transformers. Then, we try state-of-the-art models including UperNet~\cite{xiao2018unified} and Mask2Former~\cite{cheng2021mask2former}. These models are selected based on their availability of their implementations with pretrained models. More details are available in \ref{sec:seg-datails}.

\textbf{Quantitative Results}. We present the results in Table~\ref{tab:semanticseg}. For evaluation, we compute the mean Intersection-over-Union (mIoU), a widely accepted measure for assessing the performance of semantic segmentation models. Overall, SwinT-Mask2Former achieves the highest mIoU of 92.81. ConvNeXt-UperNet also has a high mIoU of 92.44 and outperforms Mask2Former on some classes, whereas Mask2Former is significantly better for vacuoles. ResNet-FCN and SegFormer exhibit lower mIoU scores of 90.78 and 91.48, respectively. The individual class IoU scores reveal that vacuoles are consistently the most challenging class to segment across all four models. For instance, ConvNeXt-UperNet achieves an IoU of 78.54 for vacuoles, while the IoUs of other classes are over 90. \revision{Additionally, Table~\ref{tab:semanticsegf1} shows the per-class F1 scores, which tell a similar story to IoU. The challenge of vacuoles can also be quantitatively confirmed by the lower F1 scores as well as by the class imbalance shown in Figure~\ref{fig:seg-distribution}.}

\begin{table}[!tb]
    \centering
    \caption{mIoU (\%) on Semantic Segmentation.}
    {\footnotesize
    \begin{tabular}{@{}lcccccc@{}}
        \toprule
        Model & Nucleus & Cytoplasm & Vacuoles & Platelets & RBC & Mean \\
        \midrule
        ResNet-FCN & 96.61 & 94.64 & 74.51 & 89.51 & 98.61 & 90.78 \\
        SegFormer & 97.18 & 95.32 & 76.74 & 89.22 & 98.97 & 91.48 \\
        SwinT-UperNet & 97.30 & 95.62 & 78.31 & 90.67 & 99.08 & 92.19 \\
        ConvNeXt-UperNet & \textbf{97.60} & \textbf{95.99} & 78.54 & 90.92 & \textbf{99.16} & 92.44 \\
        SwinT-Mask2Former & 97.55 & 95.88 & \textbf{80.03} &  \textbf{91.51} & 99.10 & \textbf{92.81} \\
        \bottomrule
    \end{tabular}%
    }
    \label{tab:semanticseg}
\end{table}

\begin{table}[!tb]
    \centering
    \caption{\revision{Per-class F1 (\%) score on Semantic Segmentation.}}
    {\footnotesize
    \revision{
    \begin{tabular}{@{}lcccccc@{}}
        \toprule
        Model & Nucleus & Cytoplasm & Vacuoles & Platelets & RBC \\
        \midrule
        ResNet-FCN & 98.28 & 97.24 & 85.40 & 94.46 & 99.30 \\
        SegFormer & 98.57 & 97.60 & 86.84 & 94.30 & 99.48 \\
        SwinT-UperNet & 98.63 & 97.76 & 87.83 & 95.11 & 99.54 \\
        ConvNeXT-UperNet & \textbf{98.79}& \textbf{97.95} & 87.98 & 95.24 & \textbf{99.58} \\
        SwinT-Mask2Former & 98.76 & 97.89 & \textbf{88.91} & \textbf{95.57} & 99.55 \\
        \bottomrule
    \end{tabular}%
    }
    }
    \label{tab:semanticsegf1}
\end{table}

\begin{figure*}[!tb]
    \centering
     \includegraphics[trim=0 0 0 0, clip, width=0.75\linewidth]{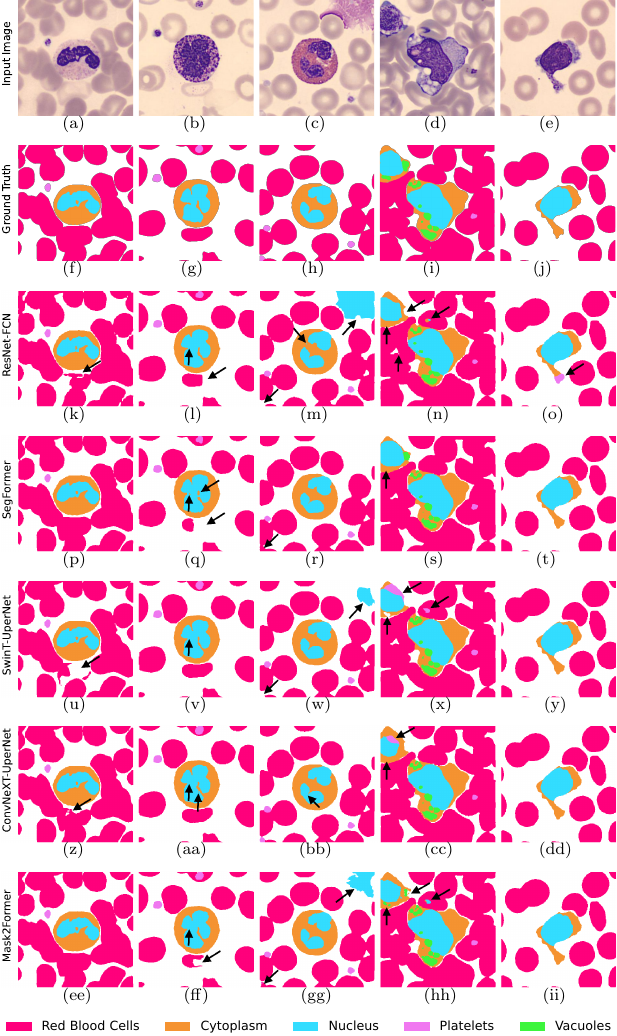}
    \caption{Segmentation results with notable errors annotated. \revision{ConvNeXt-UperNet and Mask2Former perform the best, as quantified in Table~\ref{tab:semanticseg}.}}
    \label{fig:sampleseg}
\end{figure*}

\textbf{Qualitative Results}: In Figure~\ref{fig:sampleseg}, we show five sample images (a)-(e), along with their corresponding ground truth masks (f)-(j) and masks (k)-(ii) predicted by the five models. Notable errors are marked with arrows to highlight areas of concern. For instance, red blood cells often suffer from under-segmentation, particularly when their color is pale, as demonstrated in (k), (l), (q), (u), (z), and (ff). Background debris from broken cells frequently confuses the models, leading to errors like those seen in (m), (w), and (gg). Irregular cell shapes can pose challenges for segmentation models, as observed in (o). Furthermore, vacuoles present a consistent challenge, with each model exhibiting slightly different errors, as illustrated in (n), (s), (x), (cc), and (hh).

\subsection{Attribute Recognition Without Cell Segmentation}\label{sec:attprediction}
While various applications are possible with our attributes (Sec.~\ref{sec:applications}), a preliminary question emerges: How well do standard visual recognition models perform in predicting these attributes? This is a crucial question because if the model cannot even recognize these attributes, we cannot reliably develop interpretable models on top of it. To investigate this, we conduct experiments to predict the 11 attributes from WBC images. 

\begin{figure*}[!tb]
  \centering
\includegraphics[trim=0 578pt 0 0, clip, width=0.7\linewidth]{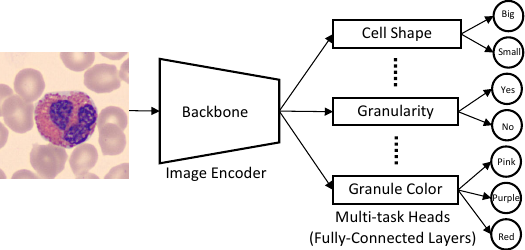}
  \caption{Baseline model architecture to predict attributes in a multi-task manner.}
  \label{fig:attpred_model}
\end{figure*}

\textbf{Baseline Models}. Our baseline model has an image encoder and multi-task head per attribute, as shown in Figure~\ref{fig:attpred_model}. We intentionally keep the model this simple and provide a foundation for future work.  We experimented VGG~\cite{vgg}, ResNet~\cite{he2016resnet}, ViT~\cite{dosovitskiy2021an}, and ConvNeXt~\cite{liu2022convnet} as backbones. The implementation details are in \ref{sec:attpred-datails}.

\textbf{Results}. We present the results in Table~\ref{tab:attpred} (the first row for each block) with four evaluation metrics. In our view, due to the imbalance of attribute values, we should care macro F1, which calculates the harmonic mean of precision and recall. Among the four backbones, the ConvNeXt model achieves the highest average macro F1 of 91.62\%, while other backbones are 0.5 to 1 point lower. We show the macro F1 per attribute on the second row of Table~\ref{tab:attribute-prediction}. Some attributes, such as granularity, granule type, and granule color, exhibit particularly high values of over 99\%. On the other hand, nucleus shape is the most challenging attribute to predict, with the lowest score of 78.14\%, which could be due to the complexity of the nucleus shape. \ref{sec:error-analysis} discusses some examples of attribute predictions. 

\begin{table*}[!tb]
    \centering
    \caption{Performance of baseline attribute predictor (Figure~\ref{fig:attpred_model}) and our cell structure-aware model (Figure~\ref{fig:partmodel}; Context Aggregation: Max Pooling) across different backbones. \revision{Results are reported as the mean over three independent runs, with 95\% confidence intervals.}}    \label{tab:backbone_method_performance}\label{tab:attpred}
    {\footnotesize
    \begin{tabular}{lcccc}
        \toprule
        Method & Accuracy & Precision & Recall & F1 \\
        \midrule
        VGG16              & $94.00_{\pm0.04}$ & $91.06_{\pm0.24}$ & $90.80_{\pm0.33}$ & $90.87_{\pm0.11}$ \\
        + Cell Structure-aware      & $94.87_{\pm0.19}$ & $92.34_{\pm0.35}$ & $92.94_{\pm0.21}$ & $92.56_{\pm0.27}$ \\
        \midrule
        ResNet50           & $94.15_{\pm0.07}$ & $91.40_{\pm0.45}$ & $91.06_{\pm0.26}$ & $91.09_{\pm0.17}$ \\
        + Cell Structure-aware      & $94.75_{\pm0.10}$ & $92.28_{\pm0.18}$ & $92.79_{\pm0.44}$ & $92.46_{\pm0.10}$ \\
        \midrule
        ViT-B/16           & $94.05_{\pm0.03}$ & $91.49_{\pm0.28}$ & $91.01_{\pm0.38}$ & $91.11_{\pm0.14}$ \\
        + Cell Structure-aware      & $94.58_{\pm0.04}$ & $92.46_{\pm0.18}$ & $92.17_{\pm0.14}$ & $92.24_{\pm0.06}$ \\
        \midrule
        ConvNeXt-T         & $94.47_{\pm0.11}$ & $91.81_{\pm0.19}$ & $91.65_{\pm0.11}$ & $91.62_{\pm0.14}$ \\
        + Cell Structure-aware      & $95.16_{\pm0.08}$ & $92.98_{\pm0.16}$ & $93.06_{\pm0.13}$ & $92.97_{\pm0.06}$ \\
        \bottomrule
    \end{tabular}
    }
\end{table*}

\begin{table*}[tb]
    \centering
    \caption{Macro F1 (\%) per Attribute. \revision{Results are reported as the mean over three independent runs, with 95\% confidence intervals.}}
    \resizebox{\textwidth}{!}{%
    \begin{tabular}{@{}lcccccccccccc@{}}
        \toprule
        Model  & Cell size & Cell shape & Nucleus shape & NC ratio & Chrmt. density & Cyto. texture & Cyto. color & \revision{Vacuoles} & Granularity & Granule color & Granule type & (Average) \\ \midrule
       Baseline w/o Seg. (Figure~\ref{fig:attpred_model}) & $83.72_{\pm0.67}$ & $91.77_{\pm0.26}$ & $78.14_{\pm1.98}$ & $96.72_{\pm0.38}$ & $86.04_{\pm0.86}$ & $94.23_{\pm0.49}$ & $88.74_{\pm0.91}$ & $90.09_{\pm0.92}$ & $99.75_{\pm0.07}$ & $99.03_{\pm0.02}$ & $99.62_{\pm0.01}$ & $91.62_{\pm0.14}$ \\
      Seg.-based Design (Figure~\ref{fig:partmodel})    & $85.14_{\pm0.22}$ & $92.44_{\pm0.17}$ & $82.53_{\pm0.74}$ & $97.08_{\pm0.20}$ & $86.22_{\pm0.90}$ & $94.66_{\pm0.54}$ & $89.87_{\pm0.13}$ & $96.85_{\pm0.18}$ & $99.57_{\pm0.06}$ & $98.98_{\pm0.09}$ & $99.47_{\pm0.03}$ & $92.98_{\pm0.12}$ \\
        \bottomrule
        \multicolumn{12}{l}{*Backbone: ConvNeXt-T; Context Aggregation: Attention} \\
    \end{tabular}%
    }\label{tab:attribute-prediction}
\end{table*}

\subsection{Attribute Recognition With Cell Segmentation}\label{sec:segrec_exp}
\subsubsection{\textbf{Peripheral Cell Classification}}
Our segmentation annotations enable us to design a cell structure-aware model (Sec.~\ref{sec:segrec_model}), which effectively utilizes segmentation masks to mimic the way clinicians recognize cell attributes. This section empirically explores the model's performance and key design decisions of its components.

\textbf{Comparison with Multiple Image Classifiers}. We evaluate the cell structure-aware model using multiple image classification architectures from Sec.~\ref{sec:attprediction}. The results in Table~\ref{tab:backbone_method_performance} show consistent improvements across multiple evaluation metrics for all the classifiers we tested. This consistent trend highlights the value of our segmentation annotations, which enable our model to enhance cell attribute prediction effectively.

\textbf{Per-attribute Results}. Table~\ref{tab:attribute-prediction} presents the per-attribute performance, highlighting notable improvements from the baselines, particularly in nucleus shape (78.14 $\rightarrow$ 82.53) and \revision{vacuoles} (90.09 $\rightarrow$ 96.85). This indicates that explicit representation of cell segmentation, including nucleus and vacuoles, helps the model recognize these attributes better than solely from unsegmented images.

\textbf{Context Aggregation Layers}. We ablate the choice of context aggregation layers as reported in Table~\ref{tab:aggregation}. Our segmentation-aware model consistently achieves an average F1 measure of over 92\%, regardless of the choice of $f_{\text{Agg}}(\cdot)$, demonstrating an improvement from the baseline of 91.62\%. Among the different aggregation mechanisms, attention yields the best scores, achieving 92.98\%. However, with the exception of concatenation, which exhibits slightly lower performance, other simpler aggregation methods such as max-pooling produce nearly equivalent results, as their confidence intervals overlap. 

\textbf{Predicted Segmentation Mask}. While our experiments use ground truth segmentation maps by default, we also evaluate our model with predicted segmentation maps generated by ConvNeXt-UperNet from Sec.~\ref{sec:segmentation}, simulating a practical scenario. As reported in the last column of Table~\ref{tab:aggregation}, although there is a slight decline in performance (e.g., from 92.98 to 92.32 with attention aggregation), the results remain higher than the baseline of 91.62.

\textbf{Ways to Use Segmentation Masks}. We also implemented a baseline that simply concatenates the segmentation mask with the input image, meaning the model takes the input of $K+3$ channels (with 3 channels from RGB) images. As shown in Table~\ref{tab:direct_concat_seg_main}, this baseline yields a higher Macro F1 score (92.32) than the baseline without segmentation (91.62) but remains lower than our cell structure-aware recognition model (92.97). 

\begin{table}[!t]
\caption{Different context aggregation $f_{\operatorname{Agg}}(\cdot)$. Baseline model without segmentation has F1 of $91.62_{\pm0.14}$. \revision{Results are reported as the mean over three independent runs, with 95\% confidence intervals.}}
\centering
{\footnotesize	
\begin{tabular}{ccccc}
\toprule
Aggregation & Trainable? &  Input-Output Shapes & F1 (GT Seg.) & F1 (Pred. Seg.) \\
\midrule
Concatenate & No & $(B, K,d) \to (B, Kd)$ & $92.64_{\pm0.08}$ & $92.04_{\pm0.12}$\\
Max Pooling & No & $ (B, K,d)  \to (B,d)$ & $92.97_{\pm0.06}$ & $92.25_{\pm0.12}$ \\
Average Pooling & No & $(B, K,d) \to  (B,d)$ & $92.79_{\pm0.03}$ & $92.23_{\pm0.14}$ \\
Weighted Average & Yes & $(B, K,d) \to (B,d) $ & $92.92_{\pm0.10}$ & $92.27_{\pm0.14}$ \\
Attention & Yes & $(B, K,d) \to (B,d)$ & $\textbf{92.98}_{\pm0.12}$ & {$\textbf{92.32}_{\pm0.19}$} \\
\bottomrule
\multicolumn{5}{c}{*Backbone: ConvNeXt-T; *B: batch size; K: \# seg. classes; d: feature dimension} \\
\end{tabular}%
}
\label{tab:aggregation}
\end{table}

\begin{table}[!t]
    \centering
    \caption{Different ways of using segmentation masks. \revision{Results are reported as the mean over three independent runs, with 95\% confidence intervals.}}\label{tab:direct_concat_seg_main}
    {\footnotesize	
    \begin{tabular}{l c}
        \toprule
        Method & Average Macro F1 (\%) \\
        \midrule
        Baseline without segmentation mask & $91.62_{\pm0.14}$ \\
        Directly combine image and segmentation mask as inputs & $92.32_{\pm0.09}$ \\
        Cell structure-aware model (ours) & $92.97_{\pm0.06}$ \\
        \bottomrule
        \multicolumn{2}{l}{*Backbone: ConvNeXt-T; Context Aggregation: Max pooling} \\
    \end{tabular}
    }
\end{table}

\begin{table}[!t]
    \centering
    \caption{Effect of segmentation masks on attribute prediction. \revision{Results are reported as the mean over three independent runs, with 95\% confidence intervals.}}
    \label{tab:ablate-seg}
    {\footnotesize
    \begin{tabular}{l c}
        \toprule
        Segmentation Mask & Average Macro F1 (\%) \\
        \midrule
        No segmentation mask & $91.62_{\pm0.14}$ \\
        \hdashline
        Binary: RBC & $91.65_{\pm0.11}$ \\
        Binary: Platelets (Plt.) & $91.97_{\pm0.22}$ \\
        Binary: Nucleus (Nuc.) & $92.02_{\pm0.17}$ \\ 
        Binary: Vacuoles (Vac.) & $92.52_{\pm0.13}$ \\
        Binary: Cytoplasm (Cyt. ) & $92.53_{\pm0.11}$ \\
        \hdashline
        2-class: (Defined in existing work) Nuc., Cyt., Others & $92.51_{\pm0.14}$  \\
        5-class: RBC, Plt., Nuc., Vac., Cyt., Others & $92.97_{\pm0.06}$ \\
        \bottomrule
        \multicolumn{2}{l}{*Backbone: ConvNeXt-T; Context Aggregation: Max Pooling} \\
    \end{tabular}
    }%
\end{table}

\textbf{Effect of Expanded Segmentation Classes}. Our dataset includes five segmentation classes, whereas the previous dataset had only two, as shown in Table~\ref{tab:dataset_info}. To understand how our additional classes enhance the recognition of cell attributes, we conducted an ablation study by preparing different types of segmentation masks: binary masks for each class, two classes as in prior work, and five classes as in our dataset. We then tested our cell structure-aware model (aggregation: max pooling). To ensure a fair comparison, we used the images from our dataset and merged some classes to replicate the prior segmentation class definitions, as explained in \ref{sec:segrec-two-class-version}. In other words, this experiment aims to investigate the improved \textit{definition} of segmentation classes over those used in prior work, excluding the factor of increased image quantity, as our dataset has more than ten times the number of images. The results are presented in Table~\ref{tab:ablate-seg}. Binary segmentation masks for vacuoles and cytoplasm achieve higher scores ($\sim$92.5) than RBC (91.65), suggesting that some segmentation classes contribute more to attribute prediction than others. Moreover, using two classes resulted in a lower score of 92.51 compared to 92.97 when using five classes. This demonstrates the benefit of our finer-grained semantic segmentation annotations compared to the previous work.

\begin{table*}[!bt]
    \centering
    \caption{Results on bone marrow cell classification. \revision{Results are reported as the mean over three independent runs, with 95\% confidence intervals.} }\label{tab:bmc_cls}
    {\footnotesize
    \begin{tabular}{l c}
        \toprule
        Method & Average Macro F1 (\%) \\
        \midrule
        Baseline without segmentation mask & $92.22_{\pm0.27}$ \\
        Our cell structure-aware model & $92.82_{\pm0.26}$ \\
        \bottomrule
        \multicolumn{2}{l}{*Backbone: ConvNeXt-T; Context Aggregation: Max pooling} \\
    \end{tabular}
    }
\end{table*}

\subsubsection{\textbf{Bone Marrow Cell Classification}}\label{sec:bmc}
To further demonstrate the value of our dataset with segmentation maps, even for images beyond peripheral blood cells, we evaluate our cell structure-aware model on bone marrow cell images from  Matek et al.~\cite{matek2021highly}. Since this dataset does not provide attribute annotations, we frame the task as predicting typical cell types. Additionally, because segmentation annotations are unavailable, we adapt our model from Sec.~\ref{sec:segmentation} to predict segmentation maps. More details about this implementation are available in \ref{sec:bmcdetails}. The results, shown in Table~\ref{tab:bmc_cls}, indicate that our model achieved an average macro F1 score of 92.82, surpassing the baseline model without segmentation masks, which achieved 92.22.

\subsubsection{\textbf{\revisionred{Robustness of Cell Structure-Aware Model}}}\label{sec:robustness_external}
\revisionred{To investigate the robustness of the cell structure-aware model under different lighting conditions, staining, and other characteristics, we train the model on our dataset and then test it on other peripheral blood cell datasets, namely RaabinWBC~\cite{kouzehkanan2022large} and the SciRep dataset~\cite{li2023deep}. Since these datasets do not have attribute annotations or segmentation masks, we evaluate the task of predicting typical cell types using the segmentation masks predicted by the model trained in Sec.~\ref{sec:segmentation}. Additional implementation details are provided in~\ref{sec:robustnessdetails}. The results are shown in Table~\ref{tab:robustness_external}. Our cell structure-aware model achieves a macro F1 score of $99.70$ on the in-domain PBC dataset, while obtaining $56.44$ and $74.20$ on the out-of-domain RaabinWBC and SciRep datasets, respectively. The performance gap between in-domain and out-of-domain evaluations indicates that there is still room for improvement. Closing this domain gap through domain adaptation and domain generalization techniques is left for future work.}

\begin{table*}[!bt]
    \centering
    \caption{\revisionred{Macro F1 (\%) of our cell structure-aware model (Backbone: ConvNeXt-T; Context Aggregation: Max pooling) trained on PBC dataset and evaluated on both the in-domain dataset and unseen out-of-domain datasets for cell classification task. Results are reported as the mean over three independent runs, with 95\% confidence intervals.}}\label{tab:robustness_external}
    \resizebox{\linewidth}{!}{%
    {\footnotesize
    \revisionred{
    \begin{tabular}{c c c}
        \toprule
        PBC (In-Domain) & RaabinWBC~\cite{kouzehkanan2022large} (Out-of-domain) & SciRep~\cite{li2023deep} (Out-of-domain) \\
        \midrule
        $99.70\pm0.08$ & 
        $56.44\pm5.83$ &
        $74.20\pm6.67$ \\ 
        \bottomrule
    \end{tabular}
    }
    }}
\end{table*}

\section{Applications}\label{sec:applications}
We believe our datasets opens the potential for new sets of applications that were not possible with the existing datasets. To demonstrate this, this section showcases specific applications using our WBCAtt+ dataset. 

\subsection{Intelligent Cell Morphology Analyzer}\label{sec:cell-morph-analyzer}
\begin{figure}[t]
  \centering
  \includegraphics[trim=0 648pt 0 0, clip, width=\linewidth]{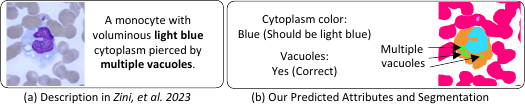}
    \caption{
     We applied the models trained on our dataset for the cell images of Covid-19 patients~\cite{zini2023coronavirus}. While not perfect, our models can recognize key features, demonstrating the applicability of our attributes. More examples are in \ref{sec:analyzer}.}
  \label{fig:broad-dataset}
\end{figure}

A practical application is to automatically analyze cell images, which provides valuable assistance to clinicians. It can also assist hematologists in searching for cell images with specific characteristics within massive datasets that would be impractical to examine manually. To this end, although we established attributes based on typical WBCs from healthy individuals, we anticipate that our attribute definitions can be applied in other contexts. For example, Zini et al.~\cite{zini2023coronavirus} discuss the morphological features of WBCs in Covid-19 patients.  We briefly inspected a small number of images reported by them. While the attributes are still applicable, we observe that the cytoplasm color, which is sensitive to staining conditions, sometimes cannot be predicted correctly due to the difference of staining.  As we can see in Figure~\ref{fig:broad-dataset} in comparison to Figure~\ref{fig:attributes}, colors look different, making the cytoplasm colors incorrectly predicted. Nonetheless, we find that attributes less sensitive of colors (e.g., \revision{vacuoles}) can be predicted correctly for the majority of the cases. We discuss this further in \ref{sec:analyzer}\revision{, including examples on other domains~\cite{alipo2022dataset,escobar2023automated, loise2022white}}.

\begin{figure}[t]
  \centering
  \includegraphics[width=0.8\linewidth]{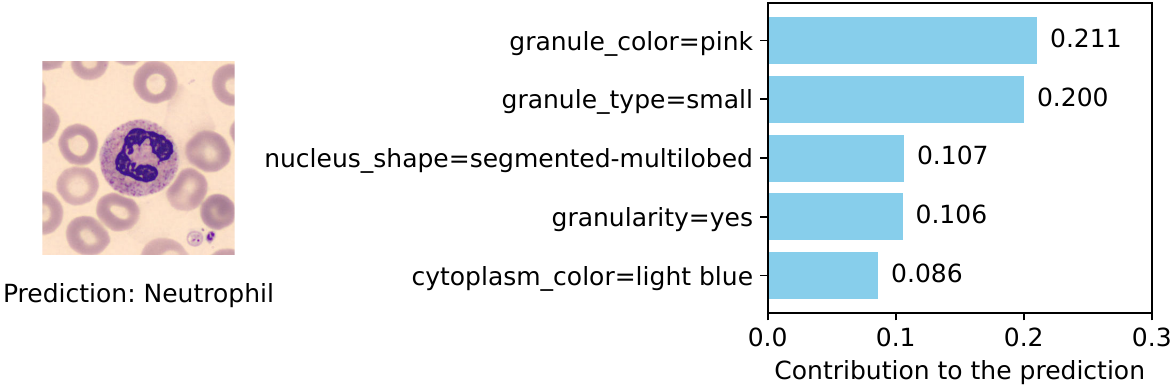}
  \caption{\revision{Top five attributes contributing to the classification of a neutrophil image by a Concept Bottleneck Model (CBM), as discussed in Sec.~\ref{sec:intervention}.}}
  \label{fig:cbm-contribution}
\end{figure}

\subsection{Human Intervention with Interpretable Models}\label{sec:intervention}
\begin{figure}[!t]
  \centering
  \includegraphics[trim=0 649pt 0 0, clip, width=\linewidth]{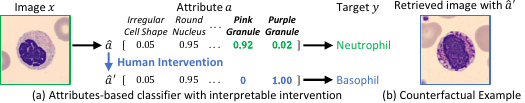}
  \caption{(a) As discussed in Sec.~\ref{sec:intervention}, our dataset enables training a cell type classifier solely based on attributes. We can ask questions like, ``What would be the predicted cell type if this cell had purple granules instead of pink?'' (b) We can retrieve corresponding images as counterfactual examples. }
  \label{fig:intervention}
\end{figure}

One of our motivations in developing this dataset is to foster XAI for WBC recognition. An effective approach to enhance explainability is designing models that make predictions based exclusively on attributes that are easily interpretable by humans. Koh et al.~\cite{koh2020concept} explored such a model that initially uses a CNN to predict a set of human-interpretable attributes and subsequently uses these attributes to predict the target output. This is called a Concept Bottleneck Model (CBM). Formally, CBMs are trained on data points of $\{ \text{image } x, \text{attribute } a, \text{target } y \}$, using $x$ to predict attributes $\hat{a}$ and then relying exclusively on $\hat{a}$ to estimate the target $\hat{y}$. We implemented a basic version of this model using the attribute predictor developed in Sec.~\ref{sec:attprediction}. In particular, we trained a L1-regularized linear softmax classifier $f(\cdot)$  to predict WBC types from the probabilities of attributes inferred from an image.  Importantly, we limit the $f(\cdot)$ to depend solely on these attribute probabilities to determine the cell category. On our dataset, the model can predict the WBC categories with a F1 of $99.40 \pm 0.04$\%, whereas a CNN predicting directly from images can achieve $99.54 \pm 0.05$\%. Despite the slightly reduced accuracy, the attribute-based model facilitates more engaging human-AI interactions, akin to Koh et al.~\cite{koh2020concept}. \revision{For example, we can investigate the top contributing attributes that drive a sample prediction by examining the weights of the linear classifier $f(\cdot)$, as illustrated in Figure~\ref{fig:cbm-contribution}.} Furthermore, the model enables human interventions by editing the attributes $\hat{a}'$ and observing how this impacts the prediction $f(\hat{a}')$ versus $f(\hat{a})$. As shown in Figure~\ref{fig:intervention}-(a), we can analyze hypothetical scenarios, such as what would happen if the cell contained granules of a different color. Such analysis would be infeasible without our dataset.

\subsection{Counterfactual Example Retrieval and Synthesis}\label{sec:counterfactual}
\begin{figure}[!t]
  \centering
  \includegraphics[trim=0 570.3pt 0 0, clip, width=\linewidth]{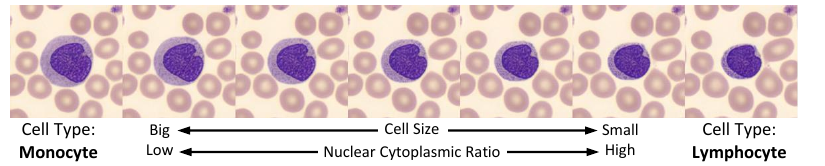}
  \caption{Our dataset enables training GANs for attribute-based image editing, which can be used for synthesizing counterfactual examples. For instance, controlling cell size and NC ratio can illustrate the decision boundaries between monocytes and lymphocytes. }
  \label{fig:attribute-edit}
\end{figure}

Another way of explaining a classifier is to show counterfactual examples, where a slight modification to the input image can result in a different classification outcome. Our attribute predictor can be utilized for retrieving counterfactual examples that correspond to attribute changes. Figure~\ref{fig:intervention}-(b) displays a retrieved example where the pink granules of the cell in Figure~\ref{fig:intervention}-(a) are changed to purple ones, illustrating the decision boundary between neutrophils and basophils. Furthermore, we can even generate counterfactual examples for a given WBC image by utilizing data-driven image editing techniques~\cite{lang2021explaining}. As a proof-of-concept, we trained an unconditional StyleGAN~\cite{Karras2019stylegan2} and implemented GAN-based editing techniques~\cite{harkonen2020ganspace,wu2021stylespace} to modify cell size and the nuclear cytoplasmic (NC) ratio with the details available in \ref{sec:gan}. The resulting images are shown in Figure~\ref{fig:attribute-edit}. These attributes are crucial in distinguishing between monocytes and lymphocytes~\cite{bain2017blood,zhang2016morphologists}. Larger cells with a lower NC ratio (depicted on the left in Figure~\ref{fig:attribute-edit}) are likely monocytes, whereas smaller cells with a higher NC ratio (shown on the right in Figure~\ref{fig:attribute-edit}) are lymphocytes. These counterfactual examples helps us understand the decision boundaries of cell classifiers.

\subsection{Discovering Dataset Biases}\label{sec:APL}
The bias issue has been reported in various domains, including face recognition models that underperform on certain genders~\cite{buolamwini18a}, races~\cite{wang2020mitigating}, or skin colors~\cite{Daneshjou2022DisparitiesID}. Similar cases have been discovered in the medical field, where skin disease detection models erroneously associate malignant cases with artificial skin markers used in clinical practices~\cite{winkler2019association}, or X-ray models accidentally learned to recognize diseases based on the appearance of a treatment device~\cite{oakden2020hidden}. Our attribute predictors can help identify these biases in cell imaging.

\begin{figure}
    \centering
  \includegraphics[width=0.65\linewidth]{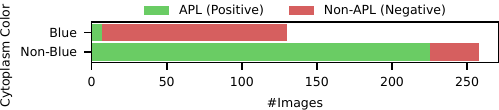}
    \caption{\revision{APL images with blue cytoplasm are significantly underrepresented compared to other categories, indicating a potential dataset bias. The figure shows the distribution of APL (positive) and Non-APL (negative) promyelocytes in the APL dataset~\cite{sidhom2021deep}, grouped by cytoplasm color predicted by our attribute prediction model.} This indicates that our model can be used to find biases in the existing cell image dataset.}\label{fig:apl}
\end{figure}

To illustrate this point, we explore the Acute Promyelocytic Leukemia (APL) dataset~\cite{sidhom2021deep}. This dataset presents a binary classification task of recognizing APL from images of immature myeloid cells, including promyelocytes -- the type of myeloid cells essential for diagnosing APL. As a preliminary step, we reproduced the baseline classifier, achieving an AUC of $76.97 \pm 2.35$\%, comparable to their~\cite{sidhom2021deep} reported $73.9$\%. Subsequently, we applied our attribute predictors, and investigated the correlation between the classifier's performance and the predicted attributes in promyelocytes. On closer inspection of all attributes, we found that promyelocytes characterized by blue cytoplasm have a much higher false negative rate (78\%) than others (57\%), which means that the model has higher chance of overlooking APL if the cells have blue cytoplasm. Examining the training set to locate the source of this bias, we found that the dataset has significantly fewer APL promyelocytes with blue cytoplasm, as shown in Figure~\ref{fig:apl}. While this could be a valid correlation if blue cytoplasm is a negative indicator of APL, our review of the literature~\cite{wang2020chromatin,sainty2000new} concludes that this is a spurious correlation specific to this dataset. Given that promyelocytes' images are \textit{not} included in our dataset, this example clearly illustrates that our attributes are versatile and help identify biases in cell imaging AI.

\subsection{Attribute-based Explainable Acute Lymphoblastic Leukemia Recognition}\label{sec:all-rec}
Although our dataset annotates morphological attributes of typical cells from healthy individuals, we believe it still offers significant potential for clinical downstream tasks, such as recognizing blood disorders. In this context, we investigate the potential use of the attributes for recognizing Acute Lymphoblastic Leukemia (ALL)~\cite{labati2011all}. As highly interpretable models, we train Concept Bottleneck Models (CBMs) (see Sec.~\ref{sec:intervention}) to classify ALL using the cell attributes. Since this dataset does not provide the attribute annotations, we use our model to predict them. As a baseline comparison, we also fine-tune a black-box end-to-end classifier of ViT-B/16. More details are available in \ref{sec:apl}.

The results are summarized in Table~\ref{tab:all-results}. The accuracy of the linear CBM model ($86.92\pm2.13$) is lower than that of the black-box prediction model ($91.28\pm2.17$). To close the gap, we trained a stronger CBM model with a gradient boosting classifier. This model achieved a comparable accuracy ($90.77\pm1.23$) with overlapping confidence intervals. This suggests the feasibility of using our attributes as interpretable features for clinical downstream tasks.

\begin{table}[!t]
\centering
\caption{Acute lymphoblastic leukemia (ALL)~\cite{labati2011all} recognition using black-box end-to-end models vs. interpretable concept bottleneck models based on the predicted cell attributes. \revision{Results are reported as the mean over three independent runs, with 95\% confidence intervals.} }
\resizebox{\linewidth}{!}{%
    \begin{tabular}{ccc}
    \toprule
    Method                                      & Interpretability        & Accuracy (\%)        \\ \midrule
    Black-box Classifier (ViT-B/16)           & -                  & $91.28_{\pm 2.17}$    \\
    Linear CBM with Predicted Cell Attributes          &   \cmark                & $86.92_{\pm 2.13}$    \\
    Gradient Boosting CBM with Predicted Cell Attributes         &  \cmark      & $90.77_{\pm 1.23}$    \\ \bottomrule
    \end{tabular}%
}
\label{tab:all-results}
\end{table}

\section{Conclusion}\label{sec:conclusion}
We have presented a densely-annotated dataset for WBC recognition, containing morphological attributes and segmentation maps for 10k cell images. This dataset addresses the current gap in the development of explainable and interpretable machine learning models for WBC analysis, a crucial task in hematology and pathology. We trained automatic attribute recognition and cell segmentation models and showcased various applications that can be developed using our dataset. We hope that our dataset will foster advancements in XAI in the fields of pathology and hematology.

\textbf{Limitation \revision{and Future Work}}.  We annotated images from a single source~\cite{acevedo2020pbc}, which is limited to typical healthy cells from peripheral blood smears. Annotating abnormal cells from blood disorders or cells from bone marrow smears, which may be of greater clinical interest, is a direction for future work. 

Our attribute definitions assume the use of the May Grünwald–Giemsa staining method, and they may appear differently with other staining methods (e.g., the cytoplasm color purple may not look purplish). Creating a similar dataset with other staining methods is future work. \revision{Consequently, the model trained on our dataset may not generalize well for some attributes when images are acquired using significantly different staining protocols, imaging magnification, or imaging hardware. A systematic study of how variations in image acquisition affect model performance is future work. In addition, these differences, or domain shift, could be mitigated by techniques such as stain adaptation~\cite{reisenbuchler2024unsupervised}, stain normalization~\cite{xu2025adaptive}, or stain deconvolution~\cite{alsubaie2017stain}, although they may also introduce some artifacts~\cite{hoque2024stain}. Investigating the effectiveness of these measures in mitigating domain shift is also future work.}

\revision{Furthermore, while our cell structure-aware model effectively leverages compositional cell parts, it currently relies on relatively simple aggregation functions. A promising future direction is to integrate graph-based representations that more explicitly align with the biological compositionality of cell morphology. For instance, treating various cell components as nodes and their spatial or functional relationships as edges using Graph Neural Networks~\cite{xupowerful2019} could provide a more robust framework for modeling complex cellular structures.}

\section*{Acknowledgments}
This project is supported by Sysmex Corporation. Computational resources are partially provided by the National Supercomputing Centre (NSCC), Singapore. The research was conducted at the Rapid-Rich Object Search Lab (ROSE) at Nanyang Technological University, Singapore.  We would like to thank Mari Kono and Joao Nunes for their insightful discussions and efforts in defining cell attributes. We thank Siyuan Yang, Takuma Yagi, Mobarakol Islam, and Lalithkumar Seenivasan for their valuable feedback on the manuscript. Additionally, we are grateful to the anonymous reviewers for their invaluable feedback and suggestions. 

\section*{Author contributions: CRediT}
 Satoshi Tsutsui: Conceptualization, Formal analysis, Investigation, Methodology, Software, Project administration, Validation, Visualization, Writing – original draft. Winnie Pang:  Conceptualization, Data curation, Investigation, Methodology, Visualization, Writing – original draft. Shuting He: Formal analysis, Software. Bihan Wen: Funding acquisition, Supervision, Writing – review and editing. 

\section*{Funding sources}
This project is supported by Sysmex Corporation. \revision{The funder has no role in restricting access to or redistribution of the dataset developed in this paper.} Computational resources are partially provided by the National Supercomputing Centre (NSCC), Singapore.

\section*{Declaration of using Generative AI and AI-assisted technologies}
The authors utilized ChatGPT to enhance the clarity and readability of the manuscript. All interactions with ChatGPT were conducted under human supervision, and the outputs were thoroughly reviewed and edited by the authors. ChatGPT was not used to generate or suggest citations.

\appendix
\section{Data Annotation}

\subsection{Attribute Definitions}\label{sec:att_def}

\subsubsection{Cell}

\textbf{Cell Size}. Cell size refers to the overall dimensions of a WBC. Generally, the size of a WBC can often be indicative of its function, maturation stage, and activation state. For differentiating between various WBC types, lymphocytes usually have a small cell size compared to other WBC types, such as monocytes and granulocytes \cite{bain2017blood}. Estimating cell size is typically done by comparing the WBC to neighboring red blood cells (RBCs; usually 6-8 $\mu$m) within the same blood sample, as RBCs provide a consistent reference for size comparison. A WBC is classified as big if its diameter is larger than twice the diameter of the RBCs. Examples of big and small cells can be found in Figure~\ref{fig:attributes}-(a)(b).

\textbf{Cell Shape}. The cell shape of WBCs is a significant morphological feature that offers insights into the cell type.
WBCs can display a range of shapes, from round to irregular, depending on the cell type and its interactions with the surrounding microenvironment, including red blood cells. Irregular shapes are more prevalent in neutrophils or monocytes due to their unique functions or maturation stages \cite{mocsai2013diverse, epelman2014origin} Additionally, WBCs can be irregular in shape due to their interactions with adjacent red blood cells. In our definition, WBCs with circular or oval shapes are categorized as round, while any other shapes are classified as irregular. Examples of round and irregular cells are shown in Figure~\ref{fig:attributes}-(c)(d).

\subsubsection{Nucleus}

\textbf{Nucleus Shape}. The shape of the nucleus can provide crucial information about the WBC types \cite{al2018classification}. Segmented nuclei are typical characteristic of neutrophils and eosinophils, with neutrophils displaying a multilobed nucleus and eosinophils often exhibiting a bilobed nucleus \cite{standring2021blood}.  Band-shaped nuclei can be found in immature neutrophils, also known as band neutrophils \cite{semmelweisband}, which are the intermediate stage in the maturation process of segmented neutrophils. Unsegmented nuclei can be observed in lymphocytes and monocytes, with lymphocytes typically having a round nucleus and monocytes having an indented or kidney-shaped nucleus \cite{bain2017blood}. An ``irregular'' class is included to accommodate the nucleus shape that does not fall under the defined classes, often for the nucleus of a basophil or monocyte. In total, we have six nucleus shapes (Figure~\ref{fig:attributes}-(i)-(n)): segmented-multilobed, segmented-bilobed, unsegmented-band, unsegmented-round, unsegmented-indented, and irregular. 

\textbf{Chromatin Density}. The density of nuclear chromatin, which refers to the compactness of chromatin within the nucleus, is an important factor for distinguishing between different types of WBCs, such as lymphocytes and monocytes. In general, lymphocytes exhibit denser, heterochromatic nuclear chromatin, while the nucleus of monocytes appears as a ``rough mesh'', which contains more loosely packed, euchromatic chromatin \cite{standring2021blood, golkaram2017role}. For the benefit of individuals who are not familiar with clinical terminology in using our annotations, we describe the chormatin density as loosely packed (euchromatic) and densely packed (heterochromatic), as illustrated in Figure~\ref{fig:attributes}-(g)(h).

\textbf{Nuclear cytoplasmic (NC) Ratio}. The NC ratio refers to the proportion of the cell's volume occupied by the nucleus relative to the cytoplasm, and can provide valuable information regarding the cell type. A WBC with a high NC ratio is typically a lymphocyte, while a lower NC ratio is characteristic of other types. In this dataset, a WBC with an NC ratio greater than 0.7 \cite{zhang2016morphologists} is considered to have a high NC ratio. This characteristic is often associated with a thin rim of cytoplasm surrounding the nucleus. Figure~\ref{fig:attributes}-(e)(f) depicts examples of WBCs with low and high NC ratios.

\subsubsection{Cytoplasm}
\textbf{Cytoplasm Color}. The color of the cytoplasm can offer valuable information regarding the cell type, as different WBCs often exhibit varying cytoplasm colors due to differences in granule content and staining affinity. Furthermore, the cytoplasmic color can offer insights into granulocyte maturation, as it tends to vary across different stages of WBC development  \cite{hema2023ouhsc}. In light of this, we have annotated the cytoplasm colors ranging from light blue to purple blue \cite{azwai2007morphological}, as shown in Figure~\ref{fig:attributes}-(q)(r)(s). %

\textbf{Cytoplasm Texture}. Cytoplasm texture also contributes to the classification of WBCs, as different cell types may exhibit unique textures due to variations in intracellular content, such as granules and other organelles. For example, tiny, dust-like purplish granules that sometimes appear in lymphocytes and monocytes give the cytoplasm a frosted or ground glass appearance \cite{walsh2004elasmobranch} while the cells lacking these dust-like granules have a clear or transparent cytoplasm, as shown in Figure~\ref{fig:attributes}-(o)(p).

\textbf{\revision{Vacuoles}}. Discrete cytoplasmic vacuoles are often found in monocytes and sometimes in neutrophils due to their phagocytosis mechanism \cite{dale2008phagocytes}, where pathogens and cell debris are engulfed and digested. The presence of these vacuoles can help identify cell types and provide insights into the cell's functional state. Vacuoles found in other cell types, such as lymphocytes \cite{van2001peripheral} and eosinophils \cite{ulukutlu1995persistent}, might be indicative of certain diseases and can aid in the diagnostic process. Examples of cells with and without \revision{vacuoles} are shown in Figure~\ref{fig:attributes}-(t)(u).

\subsubsection{Granule}
\textbf{Granularity}.  In this dataset, granularity means the presence of prominent stainable cytoplasmic granules~\cite{standring2021blood} that distinguish between granulocytes (neutrophils, eosinophils, basophils) and agranulocytes (monocytes, lymphocytes). While granules are not entirely absent in agranulocytes, they are generally found in smaller quantities and are less noticeable compared to their presence in granulocytes \cite{tigner2020histology}. Agranulocytes are categorized as having no granularity unless prominent granules are observed. Figure~\ref{fig:attributes}-(y)(z) shows a granulocyte (granularity: yes) and an agranulocyte (granularity: no).

\textbf{Granule Color}. Granule color is a distinguishing factor among granulocytes. The granules within neutrophils, eosinophils, and basophils have different colors due to their distinct compositions, which reflect their unique immune functions. Neutrophil granules are typically pink, eosinophil granules are red, and basophil granules are purple \cite{azwai2007morphological, tigner2020histology}. This attribute is particularly useful for distinguishing eosinophils from other granulocytes, as eosinophilic granules are uniquely stained red by eosin. This is due to the presence of cationic proteins within the eosinophilic granules, which bind to the eosin dye and give the cell its characteristic red color \cite{carr2021clinical}. Figure~\ref{fig:attributes}-(v)(w)(x) shows examples of them.

\textbf{Granule Type}. Granule type describes the morphological characteristics of granules in granulocytes. Neutrophils, eosinophils, and basophils each have their roles in the immune response, possessing distinct granule types filled with different substances. Neutrophils contain small, fine granules that are packed with antimicrobial proteins and enzymes, such as myeloperoxidase, lysozyme, and lactoferrin. The granules of the eosinophils are usually round and membrane-bound that compose of major basic protein. Basophils, on the other hand, possess conspicuous and coarse granules filled with histamine, heparin, and other mediators of inflammation \cite{standring2021blood}. Figure~\ref{fig:attributes}-(aa)(bb)(cc) shows examples of them.

\subsection{Attribute Annotation Process}\label{sec:annnotation}
We used Label Studio \cite{labelstudio} (see Figure \ref{fig:annotation-interface}) as the annotation tool. To ensure the accuracy and reliability of the annotations in our dataset, we implemented a comprehensive quality control process. This involved multiple steps and the participation of domain experts. The following sections outline the key aspects of our annotation quality control.

\begin{figure}[tb!]
  \centering
  \includegraphics[width=0.7\linewidth]{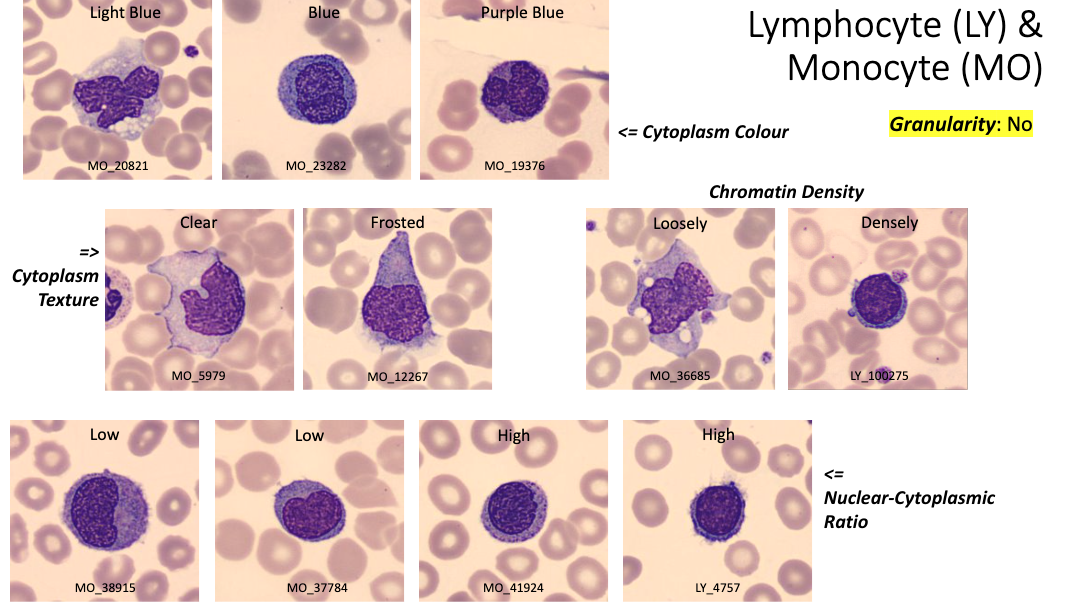}
    \caption{Example of training materials used in the annotator briefing session, illustrating visual examples of each attribute to annotate for lymphocytes and monocytes.}
  \label{fig:att_guide}
\end{figure}

(i) \textbf{Recruiting Qualified Annotators}. We called for students majoring in biomedical sciences who claimed to have a basic knowledge of WBCs. We requested each of them to provide a short description of the five types of WBCs to ensure that they possessed the required foundational knowledge. We invited those who provided satisfactory answers to an information session,  where we clearly explained the morphological attributes using materials similar to Sec.~\ref{sec:att_def} and Figure \ref{fig:attributes}. A screenshot of the material shared during this session is shown in Figure~\ref{fig:att_guide}. We then recruited them to perform a pilot annotation of 100 images (see next).

(ii) \textbf{Pilot Annotation}. After the aforementioned screening process, we instructed each candidate to independently annotate a randomly selected subset of the same 100 images. We intentionally used the same images to evaluate their understanding and enable comparisons among the candidates. We acknowledge that the candidates annotated the images while being aware of the cell type. This was a necessary compromise we made by involving medical students instead of hiring pathologists due to cost limitations. The cell types in the PBC dataset had already been verified by pathologists, and this knowledge aided the students in producing higher-quality annotations. For instance, they could utilize the cell type as prior knowledge, such as understanding that lymphocytes from healthy patients should not possess a multi-lobed nucleus.

(iii) \textbf{Annotation Phase 1 and Feedback}. Based on their performance on the pilot annotation, we selected the students and assigned them 200 images for what we refer to as Phase 1. In this phase, each student annotated a unique set of images, with allocation decisions taking into account their individual strengths. For instance, if a student demonstrated superior proficiency in annotating neutrophils compared to other cell types, we allocated a higher number of neutrophil images to that student. Subsequently, we conducted a thorough review of the annotations, providing specific feedback and updating our attribute descriptions to improve clarity.

(iv) \textbf{Annotation Phase 2 with Regular Discussions}. After completing Phase 1, we allocated the remaining images to the annotators and encouraged them to report any uncertain cases for further discussion. In cases of ambiguity, we discussed with the pathologists who defined the attributes with us, ultimately reaching a consensus. These discussions and consensus-building efforts were instrumental in ensuring the consistency and high-quality of the annotations.

(v) \textbf{Review and Validation}. Upon completion of Phase 2, our research scientists, who were not directly involved in the annotation process, meticulously reviewed and validated all annotations for correctness and adherence to the attribute definition. This means that each image in our dataset is reviewed at least by two individuals. We corrected errors or inconsistencies through this refinement process, thereby enhancing the quality and accuracy of the annotations.

(vi) \textbf{Reliability Analysis}. To assess the reliability of our annotations, we randomly selected a subset of 1,000 images, and replicated our annotation process with different annotators.  Out of the 11,000 (= 11 $\times$ 1,000) attribute annotations, 10,569 were consistent with the original annotations, resulting in an agreement rate of approximately 96.1\%. %

To appreciate and ensure high annotation quality from the annotators, we compensated them based on the number of images they labeled, with an average payment of 0.17 SGD per image.

\subsection{Segmentation Annotation Process}\label{sec:annotation-seg}
We used CVAT~\cite{boris_sekachev_2020_4009388} (see Figure~\ref{fig:annotation-interface-seg}), a web-based annotation tool, for the segmentation annotation process. We adopted a two-stage approach to improve the efficiency of the segmentation process. Initially, two research scientists carefully annotated 2000 images with the aid of the Segment Anything Model (SAM) embedded in CVAT. Subsequently, we trained a CNN segmentation model to predict segmentation maps for the remaining images. These predicted maps were then checked and refined by student annotators that are equipped with basic cell knowledge.

The CNN segmentation model used for semi-automatic annotation is UPerNet with a ConvNeXt-T backbone, implemented using the off-the-shelf MMSegmentation library~\footnote{\texttt{https://github.com/open-mmlab/mmsegmentation/tree/main/configs/convnext}}. The training data consisted of manually annotated images from the first stage, and the model was trained for 20 epochs using the AdamW optimizer with a weight decay of 0.001, a learning rate of 0.0001, and a batch size of 2.

To ensure the accuracy and reliability of the segmentation annotations, we followed a rigorous process similar to our attribute annotation procedure. We invited student annotators who had previously annotated the attributes, along with two additional students that are familiar with the attribute datasets to annotate the segmentation maps. All annotators are required to participate in a briefing session to learn about the annotation process, use the annotation tool, and establish a common baseline agreement on annotations.

\begin{figure}[tb!]
  \centering
  \includegraphics[width=0.7\linewidth]{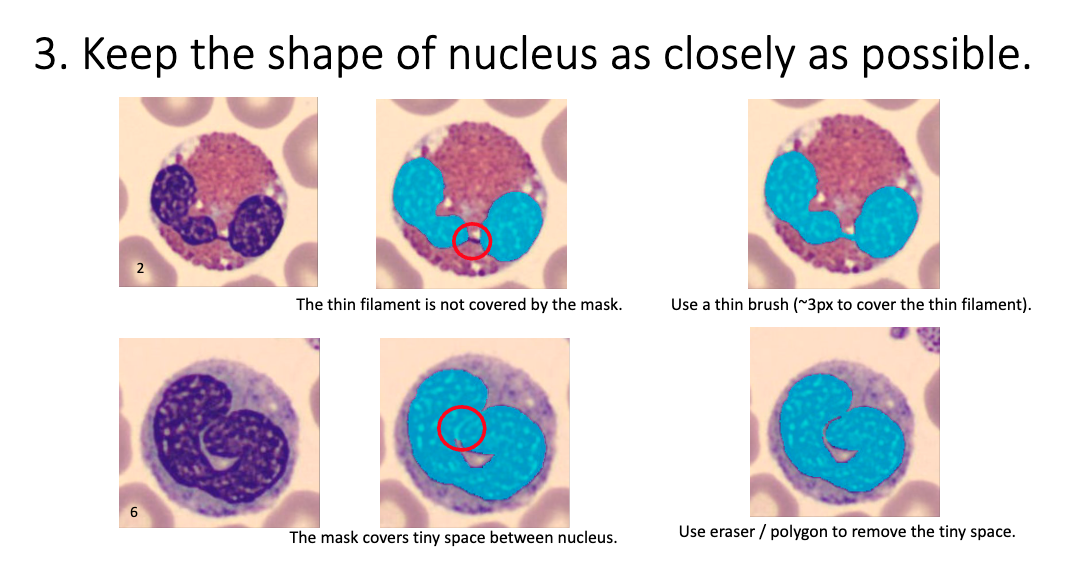}
    \caption{Guidelines for refining segmentation masks, highlighting key aspects to check and methods for refinement.}
  \label{fig:mask_guide}
\end{figure}

During the session, we emphasized the following aspects to ensure the accuracy of the maps. (A screenshot of the guideline for refining segmentation masks is provided in Figure ~\ref{fig:mask_guide}):
\begin{itemize}
    \item Presence of the nucleus, cytoplasm, and RBCs in each image.
    \item Boundaries of the WBCs and RBCs.
    \item Shape of the nucleus, particularly the filaments of segmented nuclei and nuclear indentations.
    \item Vacuoles not detected by the model.
    \item Leaked substances or debris that should not be part of the segmentation maps.
\end{itemize}

After the briefing session, each annotator was assigned the same set of 125 images (25 images per WBC class) to validate their annotation skills. Detailed and personalized feedback was provided to each annotator to ensure they met the standard. After this, the remaining images were then allocated to the annotators, and we actively engaged with them throughout the process, for the discussion on the uncertain cases. 

\begin{figure*}[tb!]
  \centering
  \includegraphics[width=\linewidth]{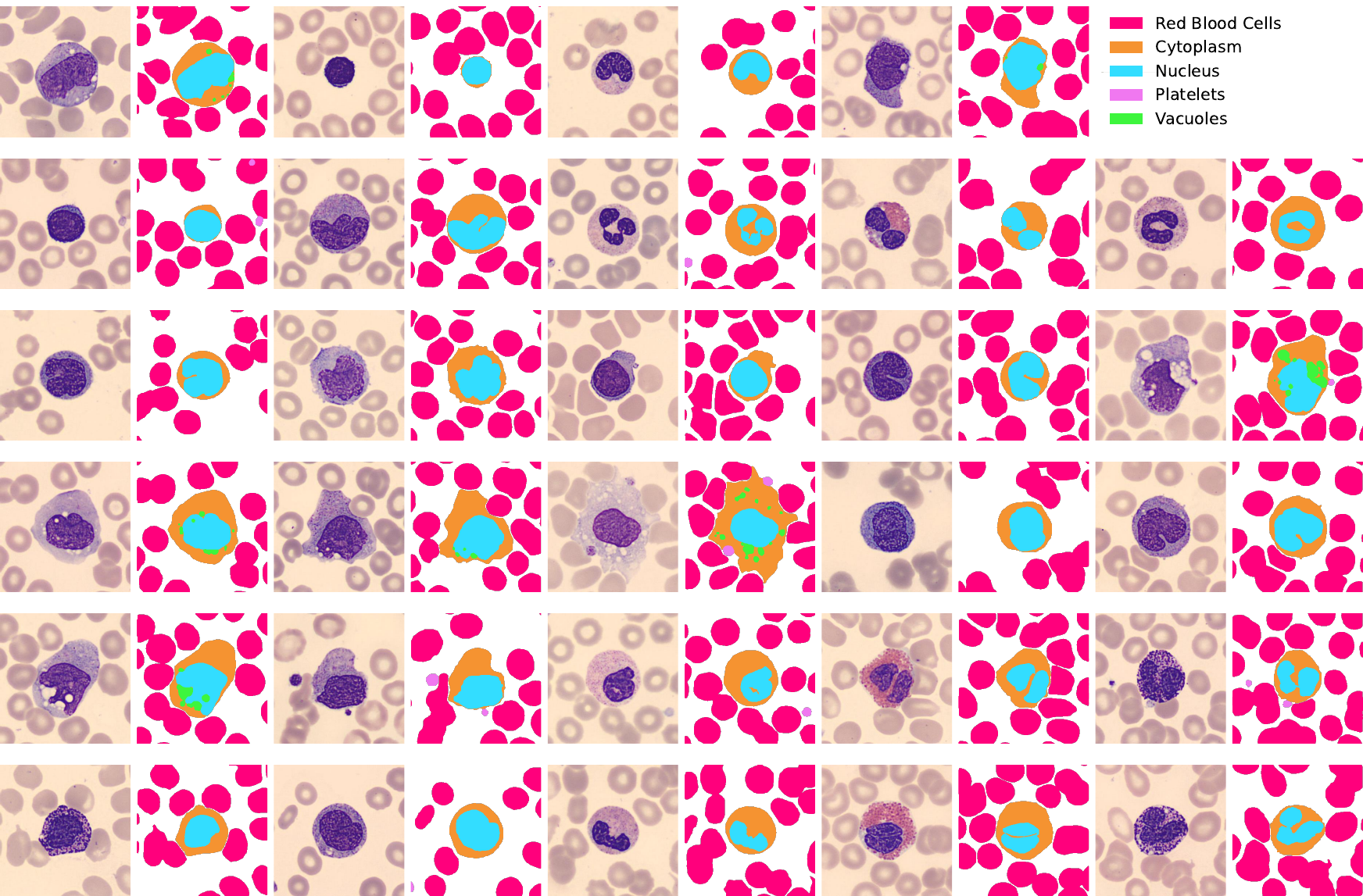}
    \caption{Examples of Annotated Segmentation Maps.}
  \label{fig:more-exp}
\end{figure*}

Once the annotations were completed, our research scientists carefully reviewed and refined each segmentation map to minimize any missed or inconsistent annotation. This process ensured that each image in our dataset was reviewed by at least two individuals. The examples of the final segmentation maps are shown in Figure~\ref{fig:more-exp}.

To appreciate and ensure high annotation quality from the annotators, we compensated them based on the number of images they refined, with an average payment of 0.2 SGD per image.

\begin{figure}[tb!]
  \centering
  \includegraphics[trim=0 546pt 0 0, clip, width=0.9\linewidth]{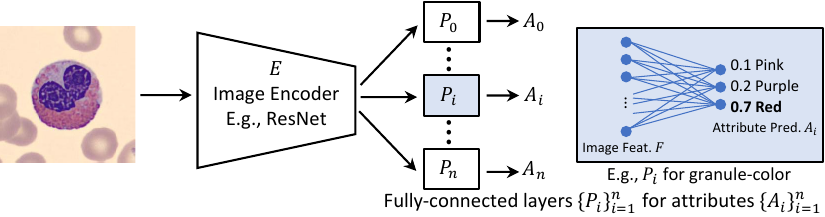}
    \caption{The architecture of the attribute prediction model. The model accepts an input image, which is processed by an image encoder $E$ to yield a feature vector. This vector is then processed by multi-task attribute predictors $\{P_i\}_{i=1}^{n}$ to produce predicted attribute distributions $\{A_i\}_{i=1}^{n}$. }
  \label{fig:network}
\end{figure}

\section{Model Details}
\subsection{Baseline Attribute Prediction Model}\label{sec:attpred-datails}
\textbf{Neural Network Architecture.}
Let $M$ denote the attribute prediction model, $I$ for the input image, and $A$ for the predicted attributes. The model $M$ consists of an image encoder $E$ and an attribute predictor $P$.

The image encoder $E$ (for example, ResNet) processes the input image $I$ and generates the feature vector $F = E(I)$.  

The attribute predictor $P$ is comprised of several attribute predictors ${P_i}_{i=1}^{n}$, each being a fully-connected layer tasked with predicting the $i$-th attribute. Each attribute predictor $P_i$ accepts the image features $F$ as input and outputs the predicted distribution $A_i$ for each attribute: $A_i = P_i(F)$. 

The predicted attributes $ A = [A_1, A_2, \ldots, A_n]$ collate all the predicted attribute distributions ${A_i}_{i=1}^{n}$, serving as the predicted attributes for the input image $I$.

The overall prediction of the attribute prediction model $M$ for the input image $I$ can be expressed as $ A = M(I) = P\left(E(I)\right), $
where $P = [P_1, P_2, \ldots, P_n]$ represents the composition of the attribute predictors.

The architecture of this model is illustrated in Figure~\ref{fig:network}. Training this network inherently involves multi-task learning, where each task is to predict a specific attribute. The model takes an input image, which is processed by a shared image encoder to yield a feature vector. This feature vector is then utilized by several attribute predictors ${P_i}_{i=1}^{n}$, each responsible for generating predicted attribute distributions ${A_i}_{i=1}^{n}$.

\textbf{Backbones.}
The backbones of the image encoder $E$ we experimented are ResNet50~\cite{he2016resnet}, VGG16~\cite{vgg}, ViT-Base/16~\cite{dosovitskiy2021an}, and ConvNeXt-T~\cite{liu2022convnet}.

\textbf{Implementation Details. }\label{sec:implementation-details}\label{sec:attpred-details}
We center-crop the image into $360 \times 360$, while the original PBC dataset provides $360 \times 362$ images. This center-cropping was not used in our conference version implementation, so the resulting numbers are slightly different, although the difference is negligible, which means that the confidence interval heavily overlaps. For training the attribute prediction model, we resize the image into $256 \times 256$ and then randomly crop it into $224 \times 224$ to feed into the image encoder. The random crop is replaced with center-crop when evaluating the model. We use the AdamW optimizer with a learning rate of 0.0001, a weight decay of 0.01, and a batch size of 96. We train the model for 2000 iterations, selecting the best-performing model based on the validation set, and evaluate its performance using the test set. Further details are available as part of the source code (\texttt{train\_att.py}). We run the code three times with three different seeds and report the 95\% confidence intervals.

\subsection{Segmentation Baselines}\label{sec:seg-datails}
We use existing segmentation models: FCN~\cite{long2015fully} based on ResNet50~\cite{he2016resnet}, SegFormer~\cite{xie2021segformer}, UperNet~\cite{xiao2018unified} based on SwinT~\cite{liu2021swin}, and ConvNeXt-T~\cite{liu2022convnet}, and Mask2Former~\cite{cheng2021mask2former} based on SwinT~\cite{liu2021swin}.

\textbf{Implementation Details}. We fine-tune the models pretrained on ADK20k~\cite{zhou2019semantic} dataset. We center-crop the image into $360 \times 360$, while the original PBC dataset provides $360 \times 362$ images. We use the AdamW optimizer with a learning rate of 0.0001, a weight decay of 0.01, and a batch size of 32. We train the model for 40 epochs, selecting the best-performing model based on the validation set, and evaluate its performance using the test set. Further details are available as part of the source code (\texttt{main\_seg.py}). Unlike attribute prediction, we run the code only once, so no confidence interval is available, following the common practice in semantic segmentation studies~\cite{long2015fully,xie2021segformer,xiao2018unified,cheng2021mask2former} (they do not report confidence intervals or error bars).

\subsection{Cell Structure-Aware Attribute Recognition Model}\label{sec:segrec-details}
\textbf{Implementation Details}. We employ the same hyperparameters and preprocessing as the baseline described in Sec.~\ref{sec:attpred-details}. We executed the code three times to obtain the mean and the 95\% confidence intervals.

\subsubsection{Minor Design Decisions}
In addition to the context aggregation mechanism, we experimented with another aspect of the model's design: the method for feeding segmented images.

In the main paper, we stated that the model receives the input of $K$ segmented images $\{I_i\}_{i=1}^{K}$ of a cell, where $I_i$ represents a cell image masked with the $i$-th semantic segmentation class. Specifically, $I_i$ has a shape of $H \times W \times 3$, where $H$ is the height, $W$ is the width, and 3 represents the RGB channels. In other words, given the original image $I$, we take the binary segmentation mask $M_i$ corresponding to the $i$-th semantic segmentation class and compute $I_i = I \odot M_i$, where $\odot$ denotes element-wise multiplication. Note that the shape of $M_i$ is $H \times W \times 1$, so we broadcast the last dimension to 3.

Although not reported in the main paper, we actually attempted an alternative approach by setting the shape of $I_i$ to $H \times W \times 4$. In this scenario, $I_i = \texttt{concat}_3(I,M_i)$, where $\texttt{concat}_3$ means the concatenation along the third dimension. This model, combined with the max-pooling aggregation, resulted in an average F1 score of $92.41\pm0.12$, which is lower than our reported model's score of $92.97\pm0.06$.

\subsection{Ablation Study w.r.t. Segmentation Classes}\label{sec:segrec-two-class-version}
Our segmentation map consists of 6 possible values: 0 for others (e.g., background, debris, etc.), 1 for cytoplasm, 2 for nucleus, 3 for platelets, 4 for RBC, and 5 for vacuoles. The existing dataset only annotates nucleus and cytoplasm, so we merge class 5 into class 1, and classes 3 and 4 into class 0.

\section{More Results}
\subsection{Baseline Attribute Prediction Model}\label{sec:error-analysis}
\textbf{Quantitative Results}. While the main paper shows aggregated results only, we present the precision, recall, and F1 score for each each attribute-value pair in the 
\texttt{./results/tab\_baseline\_attpred.md} file in the codebase. 

\textbf{Qualitative Results.} In the following subsections, we investigate specific prediction results, both correct and incorrect ones, produced by our attribute prediction model. We use Grad-CAM to highlight the areas the model focuses on. We used ResNet50 backbone. 

\subsubsection{Successful Cases on Our Dataset}
\begin{figure}[tb!]
  \centering
  \includegraphics[trim=0 434pt 0 0, clip, width=\linewidth]{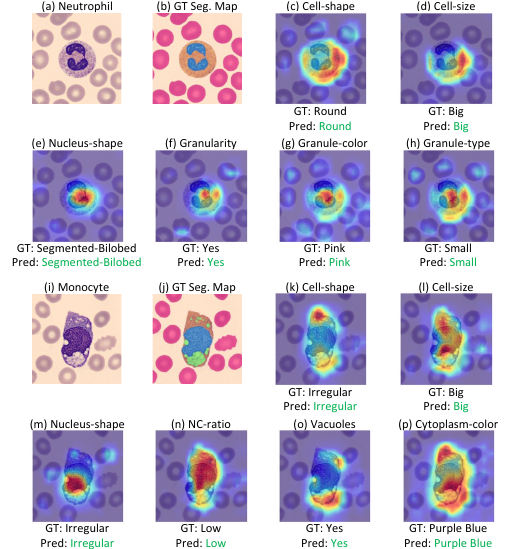}
    \caption{Grad-CAM heatmaps of accurate attribute predictions for a neutrophil ((a)-(h)) and a monocyte ((i)-(p)). Among the 11 attributes, we have selected six attributes that effectively differentiate between the respective cell types (neutrophil and monocyte).}
    \label{fig:correct-pred}
\end{figure}

When the model correctly predicted attributes, we expect that the Grad-CAM heatmap will emphasize the cell structure in line with the attribute definition. For instance, if the model is predicting the nucleus shape, it should highlight the nucleus prominently. Figure~\ref{fig:correct-pred} shows two examples of successful attribute predictions with Grad-CAM heatmaps. In these instances, the model effectively localizes the attributes. The Grad-CAM heatmaps focus on the cell edges when predicting cell shape and cell size, as shown in Figure~\ref{fig:correct-pred}-(c), (d), (k), and (l). At times, the localization is extremely precise. Particularly, for the nucleus-shape prediction in Figure~\ref{fig:correct-pred}-(e), the Grad-CAM heatmap accurately marks the thin filament of the nucleus, which serves as a key determinant for segmented nucleus identification. Moreover, the Grad-CAM can localize the \revision{vacuoles}, as shown in Figure~\ref{fig:correct-pred}-(o), and highlight the cytoplasmic area during the prediction of cytoplasm color, as depicted in Figure~\ref{fig:correct-pred}-(p).

\begin{figure*}[tb!]
  \centering
  \includegraphics[trim=0 368pt 0 0, clip, width=0.9\linewidth]{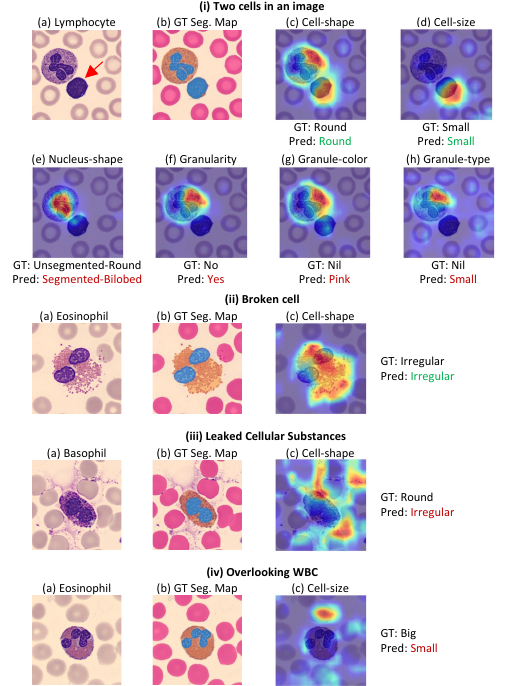}
    \caption{Potential factors contributing to incorrect attribute predictions. Please refer to \ref{sec:error-analysis} for more detailed descriptions. \textcolor{pptgreen}{Correct} predictions are highlighted in green, while \textcolor{pptred}{incorrect} predictions are highlighted in red.}    \label{fig:wrong-pred}
\end{figure*}

\subsubsection{Failure Cases on Our Dataset}
Subsequently, we investigate the cases where the model incorrectly predicted the attributes. We expect the Grad-CAM heatmaps to indicate the reasons behind the model's failures. After manually examining multiple instances of failed predictions, we have identified the following potential factors contributing to incorrect predictions:

(i) \textbf{Presence of Two Cells in an Image}. The existence of multiple distinct cell types within an image can present a challenge for the attribute predictor. Given that attributes are highly specific to different cell types, this situation can cause confusion during attribute prediction. For example, in Figure~\ref{fig:wrong-pred}-(i)(a), both a neutrophil and a lymphocyte are present in the same image. The lymphocyte is considered the ground truth for the cell type in this image, as annotated by the creator of the PBC dataset (not by us), primarily because the lymphocyte occupies a more central position. Consequently, our attribute annotations are based on the lymphocyte rather than the neutrophil. During attribute prediction, even though the Grad-CAM heatmaps focus on the lymphocyte when predicting cell-shape (Figure~\ref{fig:wrong-pred}-(i)(c)) and cell-size (Figure~\ref{fig:wrong-pred}-(i)(d)), the attribute predictor looks at the neutrophil when predicting other attributes such as granularity (Figure~\ref{fig:wrong-pred}-(i)(f)), granule color (Figure~\ref{fig:wrong-pred}-(i)(g)), and granule type (Figure~\ref{fig:wrong-pred}-(i)(h)). These attributes related to granules are typically not triggered for lymphocytes in most cases, resulting in incorrect predictions.

(ii) \textbf{Broken Cell} vs (iii) \textbf{Leaked Cellular Substances}. Sometimes the structure of cell is broken. Figure~\ref{fig:wrong-pred}-(ii) shows an broken cell where the cell membrane is unrecognizable, thus making the cell-shape irregular, which is correctly predicted. On the other hand, Figure~\ref{fig:wrong-pred}-(iii) illustrates an oval-shaped basophil surrounded by some cellular substances. Despite the presence of these substances, since its cell membrane remains clearly distinguishable, we annotated as having a round cell shape. However, the Grad-CAM heatmap reveals that the substances outside the cell are interpreted as the cell boundary, leading to an incorrect prediction of an irregular cell shape for this basophil.

(iv) \textbf{Overlooking WBC}. We have observed instances where the WBC is overlooked, and the Grad-CAM heatmap highlights the red blood cell (RBC) instead, as depicted in Figure~\ref{fig:wrong-pred}-(iv). Consequently, this leads to incorrect predictions, particularly when estimating cell size since RBCs generally have smaller sizes compared to WBCs.

\begin{table*}[tb]
    \centering
    \caption{Per-attribute Macro F1 (\%) on Attribute Prediction. Aggregation in Seg-based Model: Max Pooling. \revision{Results are reported as the mean over three independent runs, with 95\% confidence intervals.}}
    \resizebox{\textwidth}{!}{%
    \begin{tabular}{cccccccccccccc}
    \toprule
    Backbone & Model  & Cell size & Cell shape & Nucleus shape & NC ratio & Chrmt. density & Cyto. texture & Cyto. color & \revision{Vacuoles} & Granularity & Granule color & Granule type & (Average) \\
    \midrule
    ConvNeXt-T & Baseline & $83.72_{\pm0.67}$ & $91.77_{\pm0.26}$ & $78.14_{\pm1.98}$ & $96.72_{\pm0.38}$ & $86.04_{\pm0.86}$ & $94.23_{\pm0.49}$ & $88.74_{\pm0.91}$ & $90.09_{\pm0.92}$ & $99.75_{\pm0.07}$ & $99.03_{\pm0.02}$ & $99.62_{\pm0.01}$ & $91.62_{\pm0.14}$ \\
    ConvNeXt-T & Seg-based & $84.45_{\pm1.30}$ & $93.07_{\pm0.09}$ & $82.15_{\pm0.17}$ & $96.95_{\pm0.01}$ & $86.59_{\pm0.69}$ & $94.49_{\pm0.33}$ & $89.64_{\pm0.70}$ & $97.16_{\pm0.05}$ & $99.64_{\pm0.02}$ & $99.06_{\pm0.11}$ & $99.47_{\pm0.09}$ & $92.97_{\pm0.06}$ \\
    ConvNeXt-T & Two-class Seg. &  $84.33_{\pm0.26}$ & $93.09_{\pm0.43}$ & $81.85_{\pm0.25}$ & $97.10_{\pm0.10}$ & $86.54_{\pm0.46}$ & $94.57_{\pm0.56}$ & $89.86_{\pm0.67}$ & $91.10_{\pm0.85}$ & $99.65_{\pm0.02}$ & $99.04_{\pm0.06}$ & $99.46_{\pm0.06}$ & $92.51_{\pm0.14}$ \\

    ResNet50 & Baseline & $83.47_{\pm0.71}$ & $91.11_{\pm0.37}$ & $76.91_{\pm0.87}$ & $96.28_{\pm0.78}$ & $84.97_{\pm0.19}$ & $94.22_{\pm0.82}$ & $87.58_{\pm0.61}$ & $89.44_{\pm1.55}$ & $99.62_{\pm0.10}$ & $98.84_{\pm0.10}$ & $99.52_{\pm0.05}$ & $91.09_{\pm0.17}$ \\
    ResNet50 & Seg-based & $84.17_{\pm0.77}$ & $92.27_{\pm0.24}$ & $80.36_{\pm0.54}$ & $96.80_{\pm0.53}$ & $86.30_{\pm0.29}$ & $94.05_{\pm0.15}$ & $88.32_{\pm0.66}$ & $96.80_{\pm0.24}$ & $99.50_{\pm0.10}$ & $99.05_{\pm0.07}$ & $99.44_{\pm0.05}$ & $92.46_{\pm0.10}$ \\
    VGG16 & Baseline & $83.40_{\pm0.34}$ & $90.16_{\pm0.32}$ & $76.67_{\pm0.54}$ & $95.98_{\pm0.19}$ & $85.12_{\pm0.35}$ & $93.90_{\pm0.46}$ & $86.90_{\pm0.78}$ & $89.54_{\pm0.30}$ & $99.69_{\pm0.06}$ & $98.75_{\pm0.19}$ & $99.42_{\pm0.01}$ & $90.87_{\pm0.11}$ \\
    VGG16 & Seg-based & $85.04_{\pm0.21}$ & $92.23_{\pm0.44}$ & $80.16_{\pm0.99}$ & $96.51_{\pm0.34}$ & $86.54_{\pm0.46}$ & $94.32_{\pm0.14}$ & $88.33_{\pm0.99}$ & $97.11_{\pm0.13}$ & $99.62_{\pm0.04}$ & $98.96_{\pm0.04}$ & $99.37_{\pm0.09}$ & $92.56_{\pm0.27}$ \\
    ViT-B/16 & Baseline & $83.37_{\pm0.46}$ & $89.66_{\pm0.53}$ & $75.57_{\pm1.16}$ & $96.61_{\pm0.42}$ & $86.56_{\pm0.52}$ & $94.39_{\pm0.24}$ & $87.38_{\pm0.46}$ & $90.50_{\pm0.37}$ & $99.68_{\pm0.04}$ & $98.96_{\pm0.08}$ & $99.54_{\pm0.04}$ & $91.11_{\pm0.14}$ \\
    ViT-B/16 & Seg-based & $83.97_{\pm0.46}$ & $91.30_{\pm0.53}$ & $78.08_{\pm0.61}$ & $96.63_{\pm0.25}$ & $86.57_{\pm1.08}$ & $93.85_{\pm0.69}$ & $89.07_{\pm1.14}$ & $97.16_{\pm0.06}$ & $99.58_{\pm0.08}$ & $98.97_{\pm0.12}$ & $99.45_{\pm0.09}$ & $92.24_{\pm0.06}$ \\
    \bottomrule
    \end{tabular}%
    }\label{tab:attribute-prediction-all}
\end{table*}

\subsection{Cell Structure-Aware Attribute Recognition Model}\label{sec:segrec-results}

While the main paper shows aggregated results of ConvNeXt-T only, we present the F1 scores per attribute in Table~\ref{tab:attribute-prediction-all}. Across the four backbones we tried, we consistently observe improvements from the baseline. More detailed results (precision, recall, and F1 score for each attribute-value pair; models using different aggregation mechanisms) are available in \texttt{./results/tab\_segbased\_attpred.md} and \\ \texttt{./results/tab\_predsegbased\_attpred.md} in the codebase.

\section{Applications Details}

\subsection{Case Study: Cell Morphology Analyzer}~\label{sec:analyzer}
We established the attributes by analyzing five major types of WBCs derived from peripheral blood samples of healthy human individuals. Nevertheless, our attribute definitions are applicable in broader contexts. Moreover, we anticipate that the attribute predictor trained on our dataset will demonstrate generalizability to other domains in the majority of cases.  However, formally exploring this aspect requires constructing datasets and conducting rigorous evaluations, which constitute future work. As a preliminary step, we conducted small-scale case studies to evaluate the applicability of the attribute predictor to cell images beyond our dataset. Specifically, we manually examined a small number of cell images of (i) peripheral blood samples from COVID-19 patients~\cite{zini2023coronavirus}, (ii) bone marrow instead of peripheral blood~\cite{matek2021highly}, as well as (iii) peripheral blood samples from a non-human species, namely juvenile Visayan warty pigs~\cite{alipo2022dataset}\revision{, which has also been previously analyzed in prior work~\cite{escobar2023automated, loise2022white}.}

\begin{figure*}[tb!]
  \centering
  \includegraphics[trim=0 34pt 0 0, clip, width=0.85\linewidth]{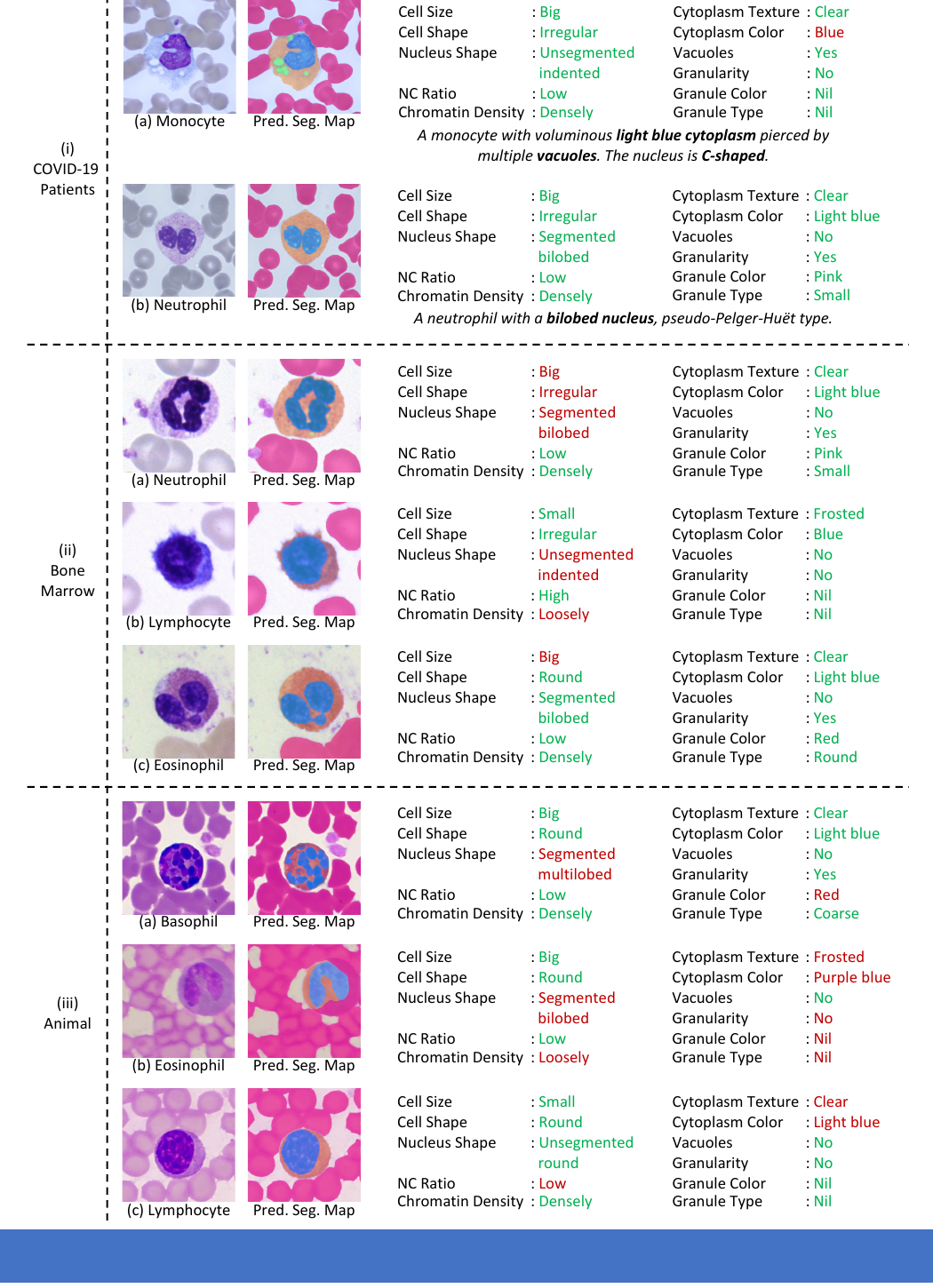}
    \caption{The prediction results of our models were evaluated for characterizing cells obtained from: (i) peripheral blood smears of patients with COVID-19, (ii) bone marrow smears, and (iii) peripheral blood smears of animal (juvenile Visayan warty pigs). The descriptions of (i)(a) and (i)(b) were summarized from \cite{zini2023coronavirus}. The predicted segmentation map (Pred. Seg. Map) is beside the input image.\textcolor{pptgreen}{Correct} attribute predictions are highlighted in green, while \textcolor{pptred}{incorrect} predictions are highlighted in red.}
  \label{fig:beyond-wbc-all-att}
\end{figure*}

(i) \textbf{Cells from COVID-19 Patients}. As shown in Figure \ref{fig:beyond-wbc-all-att}-(i), the majority of predictions made by our model, which was trained on healthy peripheral blood cells, are consistent with the descriptions provided in \cite{zini2023coronavirus} (listed at the bottom of respective image, in \textit{italic}). However, there was one exception, as the prediction related to cytoplasm color in Figure \ref{fig:beyond-wbc-all-att}-(i)(a) was inaccurately determined. This discrepancy could be attributed to the difference in staining compared to that of our training data.

(ii) \textbf{Cells from Bone Marrow}. As illustrated in Figure \ref{fig:beyond-wbc-all-att}-(ii), our attribute predictor successfully identified the attributes that related to the cytoplasm and granules, showcasing its effectiveness in investigating blood cells in bone marrow samples. However, incorrect predictions were observed for cell and nucleus-related attributes, such as cell size and nucleus shape. The poor prediction performance in cell size may be due to the nature of the bone marrow dataset, which crops the WBCs at a higher magnification level. (To observe the difference in magnification levels, compare the red blood cells in Figure \ref{fig:beyond-wbc-all-att}-(ii) to Figure \ref{fig:beyond-wbc-all-att}-(i) and Figure \ref{fig:beyond-wbc-all-att}-(iii).) Since we define the size of a cell based on its relative size compared to the red blood cells, all WBCs in Figure \ref{fig:beyond-wbc-all-att}-(ii) are small cells, even though they occupy a larger number of pixels in the image. This could also mean that our model did not actually learn the size in the way we defined it; it may be simply checking the absolute size in the image rather than checking the size relative to the red blood cells. Exploring this further is future work.

(iii) \textbf{Cells from Pigs}. To investigate the applicability beyond human blood samples, we explored animal blood smears, as shown in Figure \ref{fig:beyond-wbc-all-att}-(iii). We observe that their staining and smear preparation methods, which are different from the PBC dataset, have a substantial impact on the prediction performance, particularly for color-related attributes. In Figure \ref{fig:beyond-wbc-all-att}-(iii)(b), for instance, the model fails to predict the presence of eosinophil granules due to significant differences compared to eosinophils in other datasets. (Please refer to Figure \ref{fig:beyond-wbc-all-att}-(ii)(c) and Figure \ref{fig:wrong-pred}-(iv)(a) for the granule color of eosinophils in the bone marrow dataset and the PBC dataset, respectively.) Moreover, the nucleus and granules of Figure \ref{fig:beyond-wbc-all-att}-(iii)(a) is hardly recognizable, even to human observers. This difficulty in recognition leads to incorrect predictions for nucleus shape.

\subsection{Details of Bone Marrow Cell Classification in Sec.~\ref{sec:bmc}}\label{sec:bmcdetails}
We framed the task as predicting the five typical blood cell types from the dataset, which consists of 75,998 images. We follow an existing data split\footnote{\url{https://github.com/marrlab/DinoBloom/blob/7adf522bf3a07b990f5a151d4947e9bcf984dfc1/dinov2/eval/splits/}} to divide the dataset into 60,801 training and 15,197 testing images. To adapt the model to this dataset, we applied domain adaptation to mitigate the discrepancy between this dataset and ours, as well as incorporated global context (i.e., the entire image) into the context aggregation layer. The domain adaptation code is implemented in \texttt{main\_adapt.py}. Using a ConvNeXt-T backbone, we train the cell structure-aware model for 15,000 iterations with the AdamW optimizer (learning rate: 0.0001, weight decay: 0.01, batch size: 96) and evaluate its performance at the final iteration.

\subsection{\revisionred{Details of Robustness of Cell Structure-Aware Model in Sec.~\ref{sec:robustness_external}}} \label{sec:robustnessdetails}

\revisionred{
For PBC dataset, we follow the same dataset split as our attribute prediction dataset. For RaabinWBC and SciRep datasets, we use the same test splits provided in the code~\footnote{\url{https://github.com/PangWinnie0219/align_concept_cbm/}} of Pang et al.~\cite{pang2024integrating}. For model training, we follow the details provided in \ref{sec:bmcdetails}, except that 1) we train for only 2000 iterations because the number of images in the PBC dataset is much smaller than that of the Bone Marrow dataset, and 2) we do not use domain adaptation because the purpose of this experiment is to investigate the cross-domain robustness of the model itself without applying special measures to enhance out-of-domain performance.
}

\subsection{Details of Attribute-based WBC Classifier in Sec~\ref{sec:intervention}}\label{sec:cbmdetails}
We simply use the predicted probability (ranging from 0 to 1) of each attribute value to predict the cell types. This (lazy) formulation can cause the multicollinearity problem, so we employ the L1 regularizer. The train/val/test split is the same as our dataset split. The specific linear classifier we use is \texttt{sklearn.linear\_model.SGDClassifier( max\_iter=1000, loss="log\_loss", penalty="l1")} in scikit-learn\footnote{\url{https://scikit-learn.org/stable/modules/generated/sklearn.linear_model.SGDClassifier.html}}. 

\begin{figure}[!t]
  \centering
  \includegraphics[trim=0 572pt 0 0, clip, width=\linewidth]{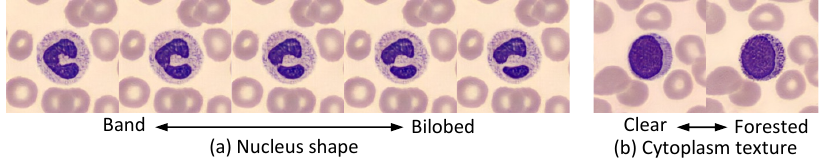}
    \caption{Applying StyleSpace~\cite{wu2021stylespace} on our attribute annotation, we discovered the way to control the necleus shape and cytoplasm-texture.}
  \label{fig:stylespace-edit}
\end{figure}

\subsection{Details of StyleGAN-based Image Editing in Sec~\ref{sec:counterfactual}}\label{sec:gan}

We trained StyleGAN-v2\footnote{\url{https://github.com/NVlabs/stylegan3}}~\cite{Karras2019stylegan2} and subsequently pixel2style2pixel (pSp)\footnote{\url{https://github.com/eladrich/pixel2style2pixel}}~\cite{richardson2021encoding} encoder.  These models enable us to embed cell images into the latent space of the trained StyleGAN, allowing for GAN inversion. Theoretically, this approach allows us to embed any cell image into the GAN latent space. By manipulating the latent space, we were able to edit the generated images, which is shown in the main paper. Specifically, we explored two techniques: GANSpace\footnote{\url{https://github.com/harskish/ganspace}}~\cite{harkonen2020ganspace}, which applies Principal Components Analysis (PCA) to the latent space, and StyleSpace\footnote{\url{https://github.com/betterze/StyleSpace}}~\cite{wu2021stylespace}, which can identify the subspace of the GAN latent space corresponding to certain attributes based on example images. Using these methods, we discovered that a specific principal component can control the cell size and NC ratio, as depicted in Figure~\ref{fig:attribute-edit} in the main paper. Additionally, although not presented in the main paper, we found subspaces to edit nucleus shape (Figure \ref{fig:stylespace-edit}-(a)) and cytoplasm texture (Figure \ref{fig:stylespace-edit}-(b)). We acknowledge that we did not rigorously evaluate controllability and image quality of these two techniques, as our primary aim was to showcase the usability of our attribute dataset. The code used for these experiments was obtained from the authors of the referenced papers (see the corresponding footnotes).

\begin{figure}[!t]
  \centering
  \includegraphics[width=0.8\linewidth]{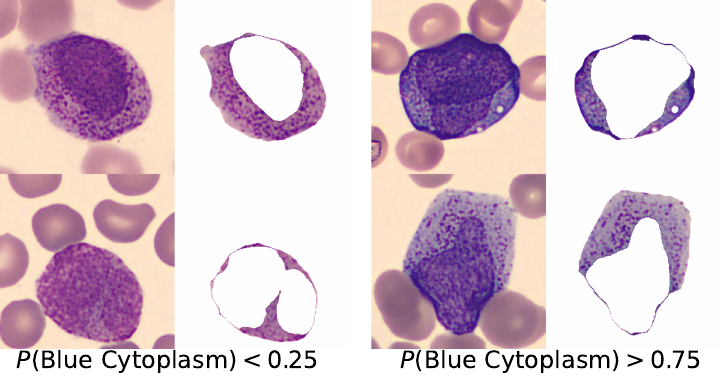}
    \caption{Cell images from the APL dataset~\cite{sidhom2021deep}. We used our model to predict the probabilities of blue cytoplasm.}
  \label{fig:blue-cyto}
\end{figure}

\subsection{Details of Dataset Bias Discovery in Sec.~\ref{sec:APL}}
We used the images provided by the dataset~\cite{sidhom2021deep} and followed their proposed data split, except that we removed the duplicated and corrupted images that we identified. We publicly reported the duplicates\footnote{\url{https://github.com/sidhomj/DeepAPL/issues/32}} and the corruptions\footnote{\url{https://github.com/sidhomj/DeepAPL/issues/31}}, but we have not received any comments thus far. Using this cleaned dataset, we reproduced the baseline classifier, achieving an AUC of $76.97 \pm 2.35$\%, comparable to their~\cite{sidhom2021deep} reported $73.9$\%. Our choice of baseline classifier is ResNet50. This classifier is used to investigate false negative rates.  

We also applied our attribute predictor to promyelocyte images. Upon manual investigation of all attributes, we found that promyelocytes characterized by blue cytoplasm have a much higher false negative rate (78\%) than others (57\%). This means that the model has a higher chance of overlooking APL if the cells have blue cytoplasm, where the example of blue cytoplasm is shown in Figure~\ref{fig:blue-cyto}.

As part of the investigation to identify the source of this bias, we discovered a bias in the training data regarding blue cytoplasm in non-APL cases, as illustrated in Figure~\ref{fig:apl} in the main paper. It is worth noting that the inherent morphological differences between APL and non-APL promyelocytes may lead to these distinct distributions. In other words, if the blue cytoplasm is a crucial feature for diagnosing APL, the bias is not necessarily undesirable. However, in this specific case, we found that the correlation only existed in the training set, and we could not find any literature supporting this bias. Consequently, we concluded that the correlation was spurious.

\subsection{Details of Attribute-based APL Recognition In Sec.~\ref{sec:all-rec}}~\label{sec:apl}
We employ the ALL-IDB~\cite{labati2011all} dataset, specifically the ALL-IDB2 subset, which includes cropped cell images. 

Due to the lack of attribute annotations for the ALL-IDB dataset, we employ our attribute predictor to obtain attributes. We utilize the ViT-B/16 backbone for this task, as our preliminary examination of a small number of cases indicates that the predicted attributes are more accurate than those from other backbones. This observation aligns with previous research~\cite{bai2021transformers}, which suggests that ViTs typically yield higher cross-dataset accuracy than CNNs. We also implement a simple domain adaptation technique to concurrently minimize the classification loss and the Maximum Mean Discrepancy (MMD)~\cite{baktashmotlagh2016distribution} loss between the labeled domain (our dataset) and the unlabeled test domain (the ALL-IDB dataset). 

The linear CBM model is the same as Sec.~\ref{sec:cbmdetails}. The gradient boosting classifier\footnote{\url{https://scikit-learn.org/stable/modules/generated/sklearn.ensemble.GradientBoostingClassifier.html}} is configured as \texttt{GradientBoostingClassifier(n\_estimators = 256, learning\_rate = 1, max\_depth = 1)}.

\section{Others}
\subsection{Peer Loss for Potential Annotation Noise}
In response to a suggestion from a reviewer in the NeurIPS submission, we have explored the peer loss~\cite{liu2020peer}, which is robust against label noise and does not require to specify the noise rate. We conducted the following experiments using the ResNet50 backbone. We executed the code three times to obtain the mean and the 95\% confidence intervals.

\begin{table}[ht]
    \centering
    \caption{Impact of Peer Loss on Average Macro F1 Score. \revision{Results are reported as the mean over three independent runs, with 95\% confidence intervals.}}
    \label{tab:peerloss}
    \resizebox{\linewidth}{!}{%
    \begin{tabular}{@{}ccc@{}}
    \toprule
    Dataset & Utilize Peer Loss? & Average Macro F1 Score (\%) \\
    \midrule
    Training Set with 10\% Label Noise & No & $85.31 \pm 0.24$ \\
    Training Set with 10\% Label Noise & Yes & $88.86 \pm 0.02$ \\
    Training Set without Added Label Noise & No & $91.20 \pm 0.06$ \\
    Training Set without Added Label Noise & Yes & $91.17 \pm 0.03$ \\
    \bottomrule
    \end{tabular}
    }%
\end{table}

To assess the efficacy of the peer loss on our dataset, we intentionally introduced noise by perturbing 10\% of the annotations within the training data. Specifically, from the pool of 67,969 categorical values (6,179 images $\times$ 11 attributes), we randomly selected 6,797 values (10\%) and replaced each with an alternative value chosen uniformly from the remaining possibilities. Using this perturbed data, we trained our model both with and without incorporating the peer loss. The outcomes, presented in the upper section of Table~\ref{tab:peerloss}, reveal that training on the noisy data alone yields an average macro F1 score of $85.31 \pm 0.24$. However, integrating the peer loss increases this score to $88.86 \pm 0.02$, demonstrating the efficacy of the peer loss.

After confirming the effectiveness of peer loss, we applied it to our training data without any artificially introduced noise. As shown in the lower part of Table~\ref{tab:peerloss}, utilizing peer loss yields an average macro F1 score of $91.17 \pm 0.03$, while training without it results in a score of $91.20 \pm 0.06$. The nearly identical scores indicate that the impact of peer loss is not significant, suggesting that the presence of annotation noise within our dataset is limited.

\begin{figure}[tb!]
  \centering
  \rebuttal{
  \includegraphics[trim=0 445pt 0 0, clip, width=1\linewidth]{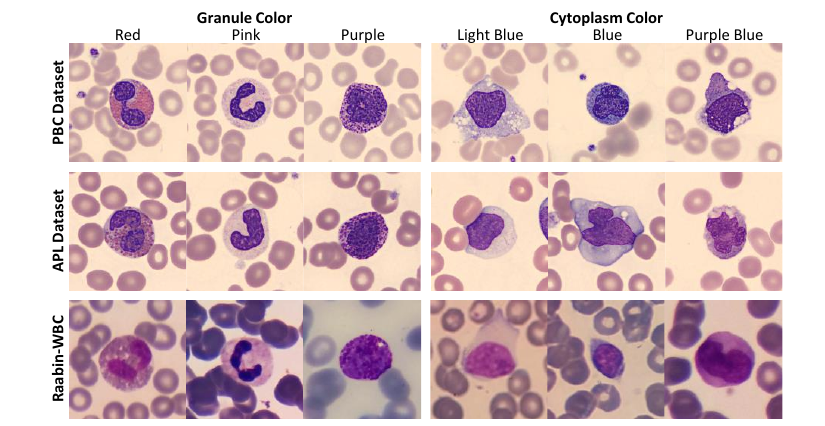}
    \caption{WBC appearances under different staining and microscopy conditions. WBCs sourced from the PBC Dataset \cite{acevedo2020pbc} (used in this work) have undergone May Grünwald-Giemsa staining and were acquired through the automated digital cell morphology analyzer, CellaVision DM96. WBCs from the APL Dataset \cite{sidhom2021deep} are stained using Wright's method and observed through the CellaVision DM100 cell analyzer. On the other hand, the RaabinWBC \cite{kouzehkanan2022large} images, stained with Giemsa, are captured using smartphones mounted on both Olympus CX18 and Zeiss microscopes. }
  \label{fig:other-stainings}
  }
\end{figure}

\begin{figure}[b!]
  \centering
  \rebuttal{
  \includegraphics[width=\linewidth]{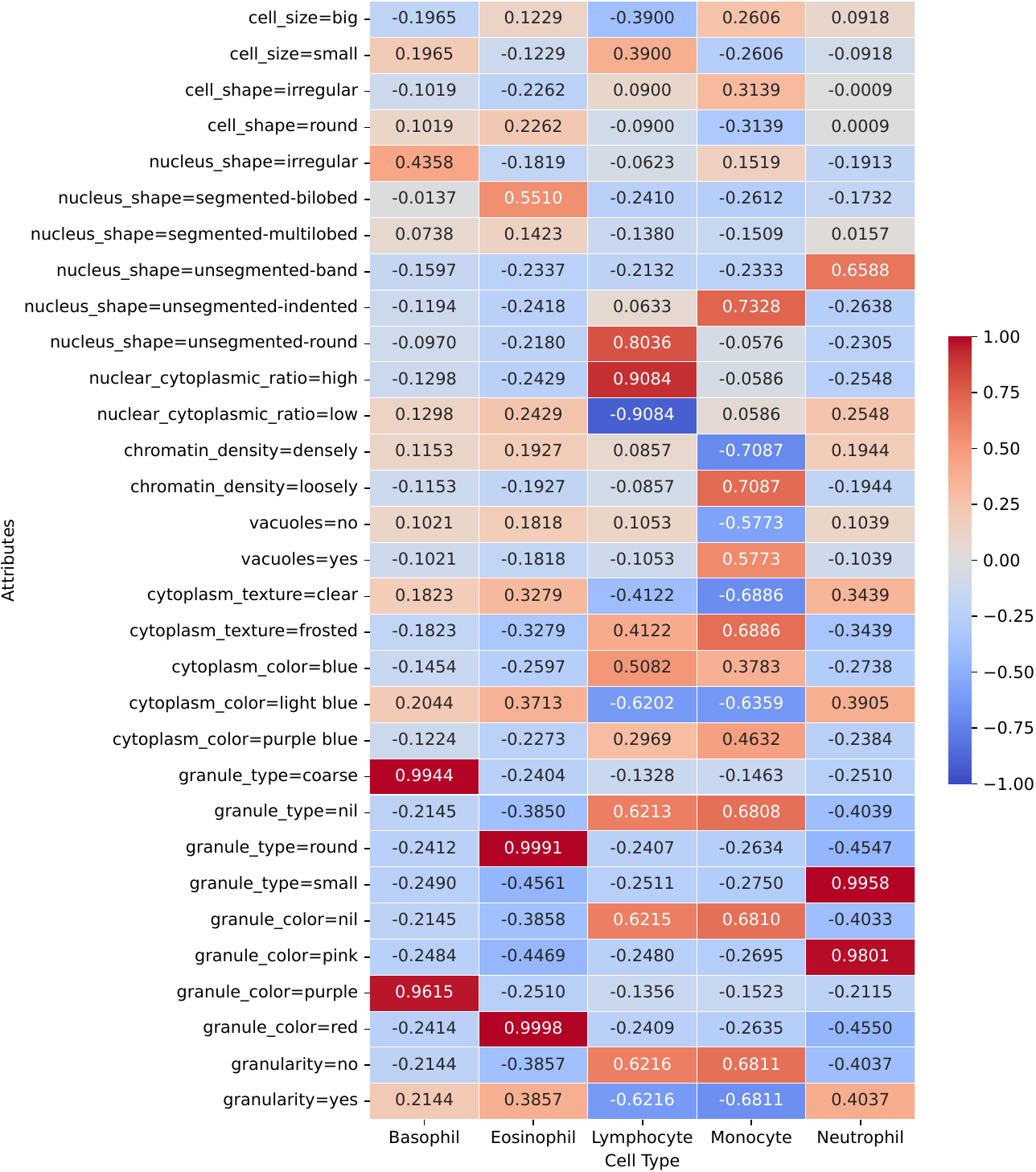}
    \caption{\revision{Correlation between binarized morphological attributes and WBC classes, computed from ground truth labels. }}
  \label{fig:att-correlation}
  }
\end{figure}

\begin{table}[t!]
    \caption{Attribute Dist. The distribution represents the results of annotating all typical WBCs from the PBC dataset, which is the image source we utilized. We did not actively control or manipulate the distribution.}
    \label{tab:attributes}
    \centering
    {\footnotesize
    \begin{tabularx}{\linewidth}{lX}
    \toprule
    Attribute & Value (Count) \\
    \midrule
    Cell-Size & Big (4,997), Small (4,271) \\
    Cell-Shape & Round (7,173), Irregular (2,095) \\
    Nucleus-Shape & Segmented-Bilobed (2,806), Unsegmented-Band (2,356), Unsegmented-Indented (1,205), Segmented-Multilobed (1,143), Unsegmented-Round (967), Irregular (791) \\
    Nuclear-Cytoplasmic-Ratio & Low (8,148), High (1,120) \\
    Chromatin-Density & Densely (8,443), Loosely (825) \\
    \revision{Vacuoles} & No (8,559), Yes (709) \\
    Cytoplasm-Texture & Clear (7,429), Frosted (1,839) \\
    Cytoplasm-Color & Light Blue (7,011), Blue (1,273), Purple Blue (984) \\
    Granule-Type & Small (3,003), Round (2,801), Nil (2,374), Coarse (1,090) \\
    Granule-Color & Pink (2,925), Red (2,803), Nil (2,373), Purple (1,167) \\
    Granularity & Yes (6,896), No (2,372) \\
    \bottomrule
    \end{tabularx}
    }
\end{table}

\begin{table}[t!]
    \centering
    \caption{Coarse Morphological Attributes.}
    \resizebox{\linewidth}{!}{%
    \begin{tabular}{@{}cccccc@{}}
        \toprule
        Cell Type & Nucleus Structure & NC Ratio & Granularity & Granularity Color & Cell Size \\
        \midrule
        Basophils & Segmented & Low & Yes & Blue / Black (dense) &  \\
        Eosinophils & Segmented & Low & Yes & Red &  \\
        Lymphocytes & Unsegmented & High & No &  & Small \\
        Monocytes & Unsegmented & Low & No &  &  \\
        Neutrophils & Segmented & Low & Yes & Blue &  \\
        \bottomrule
    \end{tabular}
    }%
    \label{tab:initial-att}
\end{table}

\begin{table}[t!]
    \centering
    \caption{\revision{Summary of updates from the conference version~\cite{tsutsui2023wbcatt}.}}
    \label{tab:journal-update}
    \resizebox{\linewidth}{!}{%
    \revision{
    \begin{tabular}{lcc}
    \toprule
     & \begin{tabular}[c]{@{}c@{}}Conference Version\\ (NeurIPS 2023)\end{tabular} & \begin{tabular}[c]{@{}c@{}}Jornal Extension\\ (Our Manuscript)\end{tabular} \\ \midrule
    \textbf{Dataset Proposed} & \textbf{WBCAtt} & \textbf{WBCAtt+} \\ \midrule
    \textbf{Attribute Annotation} &  & Sec.~\ref{sec:attribute-annotation-main} \\
    Number of Images & 10,298 & 10,298 \\
    Number of Attribute & 11 & 11 \\  \midrule
    \textbf{Segmentation Annotation} &  & Sec.~\ref{sec:seg-anno}  \\
    Number of Images & - & 10,298 \\
    Number of Segmentation Class & - & 5 \\ \midrule
    \textbf{Model Architecture} &  & Sec.~\ref{sec:segrec_model} \\
    Cell Structure-Aware Model & - & \checkmark \\  \midrule
    \textbf{Experiments} &  &  Sec.~\ref{sec:exp}\\
    Cell Segmentation & - & \checkmark  \\
    Attribute Recognition w/o Segmentation & \checkmark & \checkmark  \\
    Attribute Recognition with Segmentation & - & \checkmark \\ \midrule
    \textbf{Applications} &  &  Sec.~\ref{sec:applications}\\
    Intelligent Cell Morphology Analyzer & \checkmark & \checkmark \\
    Human Intervention with Interpretable Models & \checkmark & \checkmark \\
    Counterfactual Example Retrieval and Synthesis & \checkmark & \checkmark \\
    Discovering Dataset Biases & \checkmark & \checkmark \\
    \begin{tabular}[c]{@{}l@{}}Attribute-based Explainable \\ Acute Lymphoblastic Leukemia Recognition\end{tabular} & \checkmark & \checkmark  \\
    \bottomrule
    \end{tabular}}
    }
\end{table}

\clearpage
\subsection{Other tables and figures}
\begin{itemize}
    \item Table~\ref{tab:attributes}: Frequencies of attribute values, illustrating content similar to Figure~\ref{fig:distribution} in the main paper.
    \item Table~\ref{tab:initial-att}: Initial attributes used as a starting point to differentiate the five types of WBCs.
    \item Figure~\ref{fig:other-stainings}: Visualization of how staining conditions and imaging equipment influence the attributes. In Sec.~\ref{sec:conclusion}, we stated that our attribute definitions assume the use of the May Grünwald-Giemsa staining method, and they may appear differently with other staining methods. This figure shows examples of how the colors differ depending on the staining.
    \item \revision{Figure~\ref{fig:att-correlation}: Each cell shows the Pearson correlation coefficient (or phi coefficient for binary variables) between an attribute and a WBC type.}
    \item \revision{Table~\ref{tab:journal-update}: Summary of the differences between this manuscript and our conference version~\cite{tsutsui2023wbcatt}.}

\end{itemize}

\clearpage
\bibliographystyle{elsarticle-num}
\bibliography{references}

\end{document}